\documentclass[lettersize,journal]{IEEEtran}
\usepackage{amsmath,amsfonts}
\usepackage{algorithm}
\usepackage{array}
\usepackage{subfigure}

\usepackage{textcomp}
\usepackage{stfloats}
\usepackage{threeparttable}
\usepackage{url}
\usepackage{verbatim}
\usepackage{graphicx}
\usepackage{cite}
\usepackage{diagbox}
\usepackage{bm}
\usepackage{bbm}
\usepackage{soul}
\usepackage{amsfonts}
\usepackage[pagebackref,breaklinks,colorlinks]{hyperref}
\usepackage{multirow}
\usepackage{algpseudocode}
\usepackage{makecell}
\usepackage{booktabs}
\usepackage{xcolor}
\usepackage{booktabs}
\usepackage{bbding}
\usepackage[top=0.7in,bottom=0.5in,left=0.4in,textwidth=7.7in]{geometry}

\begin{document}
	
	\title{Scattering Model Guided Adversarial Examples for SAR Target Recognition: Attack and Defense}
	
	\author{\IEEEauthorblockN{Bowen Peng, Bo Peng, Jie Zhou, Jianyue Xie, and Li Liu, \emph{Senior Member, IEEE}}
		\thanks{Bowen Peng, Bo Peng, Jie Zhou, and Jianyue Xie are with the College of Electronic Science, National University of Defense Technology (NUDT), Changsha 410073, China
			(\url{pbow16@nudt.edu.cn}, \url{ppbbo@nudt.edu.cn}, \url{zhoujie_@nudt.edu.cn}, \url{xiejianyue0125@163.com}).}
		\thanks{Li Liu is with the College of System Engineering, NUDT, Changsha, 410073, China (\url{dreamliu2010@gmail.com}).}
		\thanks{Corresponding author: Bo Peng and Li Liu.}}
	
	\markboth{Submitted to: IEEE Transactions on Geosicence and Remote Sensing}%
	{Shell \MakeLowercase{\textit{et al.}}: A Sample Article Using IEEEtran.cls for IEEE Journals}

	\maketitle
	
	\begin{abstract}
		Deep Neural Networks (DNNs) based Synthetic Aperture Radar (SAR) Automatic Target Recognition (ATR) systems have shown to be highly vulnerable to adversarial perturbations that are deliberately designed yet almost imperceptible but can bias DNN inference when added to targeted objects. This leads to serious safety concerns when applying DNNs to high-stake SAR ATR applications. Therefore, enhancing the adversarial robustness of DNNs  is essential for implementing DNNs to modern real-world SAR ATR systems. Toward building more robust DNN-based SAR ATR models, this article explores the domain knowledge of SAR imaging process and proposes a novel Scattering Model Guided Adversarial Attack (SMGAA) algorithm which can generate adversarial perturbations in the form of electromagnetic scattering response (called adversarial scatterers). The proposed SMGAA consists of two parts: 1) a parametric scattering model and corresponding imaging method and 2) a customized gradient-based optimization algorithm. First, we introduce the effective Attributed Scattering Center Model (ASCM) and a general imaging method to describe the scattering behavior of typical geometric structures in the SAR imaging process. By further devising several strategies to take the domain knowledge of SAR target images into account and relax the greedy search procedure, the proposed method does not need to be prudentially finetuned, but can efficiently to find the effective ASCM parameters to fool the SAR classifiers and facilitate the robust model training. Comprehensive evaluations on the MSTAR dataset show that the adversarial scatterers generated by SMGAA are more robust to perturbations and transformations in the SAR processing chain than the currently studied attacks, and are effective to construct a defensive model against the malicious scatterers.
	\end{abstract}
	
	
	\section{Introduction}
	\IEEEPARstart{S}{ynthetic} Aperture Radar Automatic Target Recognition (SAR ATR) is a fundamental and challenging problem in SAR image interpretation, and has various applications in both civil and military fields \cite{zhu2021deep,yue2020synthetic,kechagias2021automatic}.
	Therefore, for the past several decades SAR ATR has received significant attention.	Recently, Deep Neural Networks (DNNs) have brought considerable progress in remote sensing image analysis \cite{zhu2017deep,liu2020deep,liu2019texture}, but are largely limited to the analysis of optical imagery. Certainly, DNNs have been introduced in SAR image interpretation tasks including SAR ATR \cite{zhang2021domain,li2021multiscale,zhao2022refine}, and have also achieved promising progress. It is reasonable to expect that the huge potential of DNNs for SAR image interpretation will be unlocked in the future \cite{zhu2017deep}.

	Due to the high-stakes nature of military and homeland security applications, SAR ATR techniques need to be highly reliable and safe, in addition to being accurate. However, there are many challenges leading to safety concerns of DNNs when used in many high-stakes applications like SAR ATR. A critical problem is that DNNs are shown to be highly vulnerable to adversarial examples, \emph{i.e.}, images added with intentionally designed yet imperceptible perturbations that are usually unnoticed to human visual system but can fool DNNs \cite{serban2020adversarial}. For instance, aircraft could be maliciously fooled as birds via adversarial attack.	There has been a great amount of work finding \cite{Duan_2021_CVPR,Rampini_2021_CVPR}, understanding \cite{fawzi2017robustness,NEURIPS2019_e2c420d9}, and defending  \emph{digital} and  \emph{physical} adversarial examples\footnote[1]{With the online API, \emph{digital} adversarial examples can be directly fed into the victim DNN. In contrast, \emph{physical} attacks design the adversarial examples that remain highly aggressive after being actively imaged by the victim's own sensing system.} \cite{madry2018towards,mart}. There exists	an attack$-$defense race in the deep learning community, and it has been demonstrated that guaranteeing  adversarial robustness is a big challenge \cite{HUANG2020100270}.

	Research on adversarial attacking and defending DNN-based systems has also drawn significant  attention in the remote sensing community. However, most of existing studies along this thread are also limited to optical images \cite{HUANG2020100270,geng2021deep,rsaa2,aars1} with only a few initial attempts on SAR images \cite{sarAAempirical2021,sarAAexperience2020,fastcw}. For example, \cite{sarAAempirical2021} and \cite{sarAAexperience2020} carried out a comprehensive evaluation of the adversarial vulnerability in the SAR scene classification and target recognition tasks utilizing the existing digital attack algorithms such as the Fast Gradient Sign Method  \cite{harnessing2015goodfellow} (FGSM), DeepFool \cite{moosavi2016deepfool}, and Carlini and Wagner \cite{carlini2017towards} (C$\&$W). More recently, in \cite{fastcw} the authors designed an accelerated C$\&$W algorithm for SAR ATR to pursue a balance between time consumption and attack ability. In these studies, DNNs based SAR ATR has been reported as extremely vulnerable in the face of extremely small perturbations. 
	
\begin{figure}[tbp]
	\centering
	\subfigure[Currently studied adversarial examples and the corresponding perturbations crafted by FGSM \cite{harnessing2015goodfellow} ($l_{\infty}$), DeepFool \cite{moosavi2016deepfool}, C$\&$W \cite{carlini2017towards} ($l_{2}$), and Sparse-RS \cite{croce2020sparse} ($l_{0}$). Please notice that the imperceptible perturbations are enlarged for observation, and in this article, the Parula color map is utilized for visualizing the gray-scale SAR images. \label{fig1a}]{\includegraphics[width=0.95\linewidth]{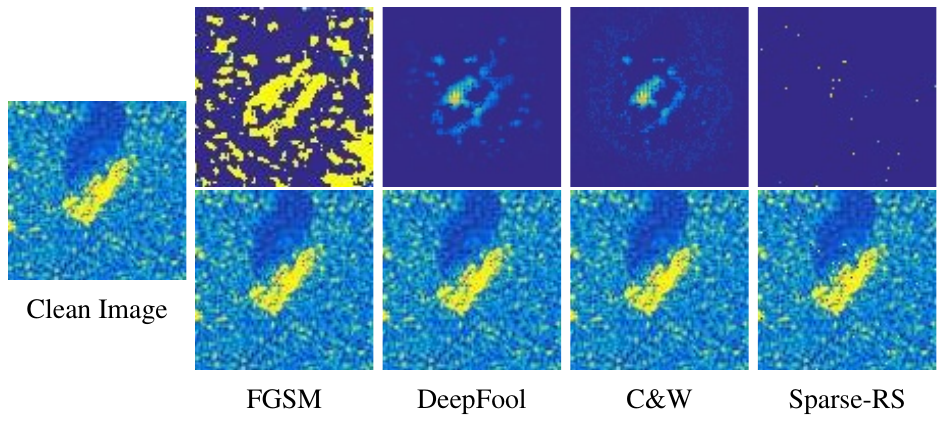} }\\
	
	\subfigure[Comparison of the pixel perturbations and scattering responses. The middle image is the distribution that contains most of energy of all the scatterers. The last two images are the adversarial scatterers and perturbed image generated by the proposed SMGAA.  \label{fig1b}]{\includegraphics[width=0.95\linewidth]{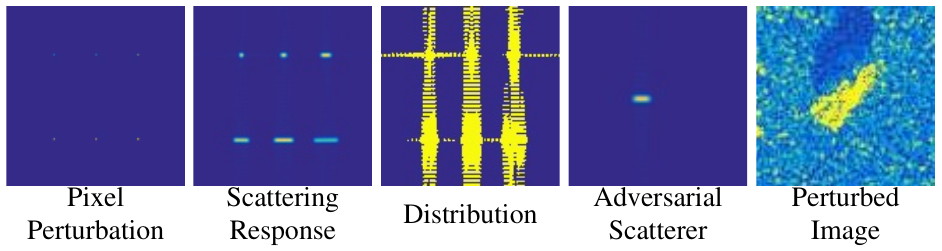} }
	\caption{Comparisons of the different types of adversarial examples.}
	\label{fig1}
\end{figure}

	However, to our best knowledge, the aforementioned works just directly transfer adversarial attacks  against optical imagery to SAR imagery, and have serious limitations. On the practical side of the research, digital attacks need to be realized physically. Then, the question of  applicability of the adversarial perturbations against physical SAR ATR systems should be given special attention as  it is significantly different from generating adversarial perturbations for natural images. Physical attacks  against optical imagery have mostly been studied in the problems of facial recognition, natural object recognition, \textit{etc}, where the natural images are usually captured at small distances such as within the sensing range of a camera on in an autonomous car or a intelligent video surveillance system \cite{Duan_2021_CVPR,zhou2018invisible}. By contrast, the applicability of physical attacks against SAR ATR systems is challenged by many factors such as prior information of the targeted SAR ATR  systems (like platform, resolution, or imaging mode), the domain knowledge of SAR image acquisition process such as the geometric scattering mechanism, where perturbations can be realized in the signal processing chain, how to make the adversarial perturbations robust to real-world perturbations inherent in the sensing chain like imaging conditions, imaging viewpoint, post processing, \textit{etc}.		For instance, as contrasted in Fig. \ref{fig1}, the scattering responses of the SAR target are intrinsically different from the currently studied perturbations, and exhibit side-lobe diffusion due to the limitations of the SAR imaging mechanism \cite{sarbook}. For another instance, 		perturbations with very small bound are sensitive to real-world perturbations and transformations \cite{guo2018countering,Xu_2018,smoothing}, \textit{e.g.}, random noise, filtering, denoising, and so on.
	
\begin{figure}[tbp]
		\centering
		\subfigure[The pipeline of the proposed SMGAA with a Trihedral adversarial scatterer was generated.]{\includegraphics[width=0.9\linewidth]{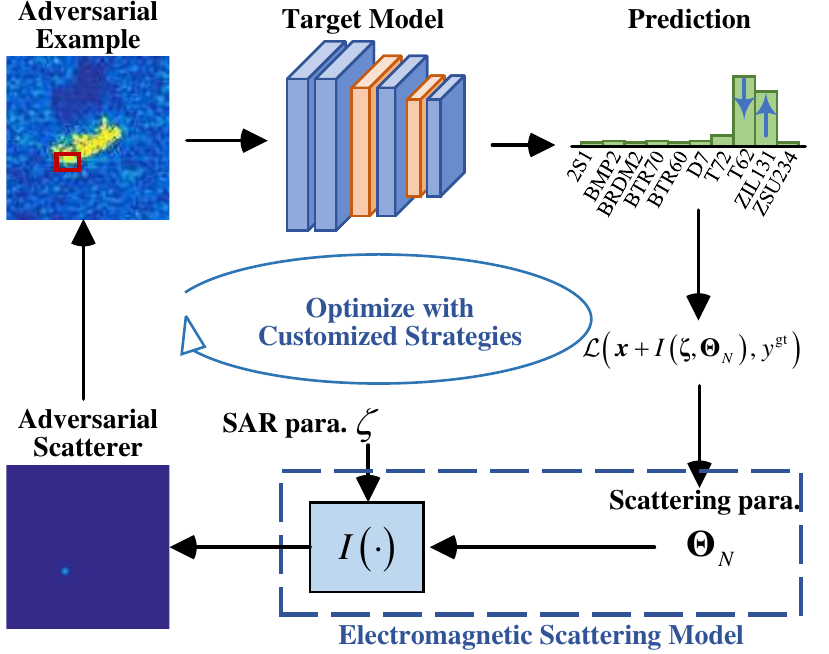}}\\	\subfigure[A toy example of describing the physical attack on a T62 tank.]{\includegraphics[width=0.9\linewidth]{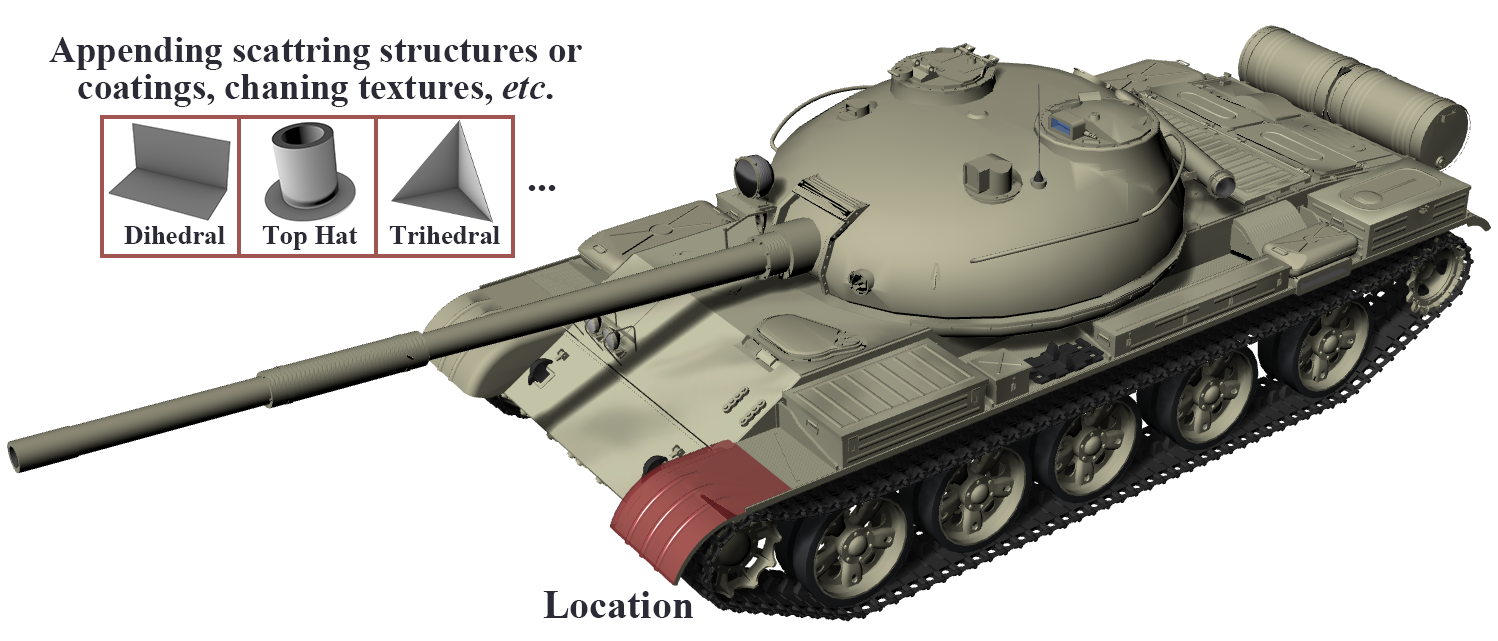}}
		\caption{The framework of the SMGAA and a toy example correlates the digital adversarial scatter with the physical scatterer deployment.}
		\label{frame}
	\end{figure}

	In this article, a novel electromagnetic-relevant attack framework, namely, Scattering Model Guided Adversarial Attack (SMGAA), is proposed to better highlight and further prevent the adversarial risks in real-world circumstances. The SMGAA aims at extending the optimization chain between perturbation image and model output to electromagnetic objects, and eventually generate more feasible and robust perturbations. Fig. \ref{fig1}-(b) demonstrates an example of SMGAA-generated perturbations which makes a well-trained AConvNet \cite{aconvnet2016chen} misclassify the rocket launcher as a tank with confidence score up to $97.43\%$. These perturbations are strictly generated by the parametric scattering model at the signal level, carry clear electromagnetic attributes, and thus are termed adversarial scatterers in this article. The illustration of the proposed framework can be found in Fig \ref{frame}.	With the Attributed Scattering Center Model (ASCM) \cite{gerry1999parametric}, a well-defined scattering model which provides concise and physically relevant attributes for both localized and distributed scattering mechanisms, we devise a customized optimization process to search for the most effective adversarial scatterers in a softened greedy manner. Firstly, the candidate scatterers will be initialized at the target and shadow region since it contains most of the structural information of the SAR target. Secondly, with the randomness and progressive adaption provided by a Gaussian stepsize generator, the optimization can escape the local optima, as well as utilize the memorization of previous steps to be more efficient and effective. The failed updates will also be adopted with a certain probability during the process to alleviate the overfitting of greedy search. Furthermore, inspired by the random restart strategy and population-based algorithms, we speed up the entire process via a population-style generation which also supports the robust mini-batch training procedure.
	
	The main contributions of this article are summarized as follows.

	\begin{enumerate}
		\item We propose a novel Scattering Model Guided Adversarial Attack (SMGAA) framework for DNN-based SAR ATR, which leverages the geometric scattering mechanism to generate parametric adversarial scatterers with well-defined physical attributes. It provides high feasibility for studying the physical adversarial threats. Moreover, it can directly serve the defense against malicious scatterers with limited additional computational cost.
		
		\item We devise a customized optimization process to exploit the most effective ASCM parameters to fool the ATR models and facilitate the robust model training. The optimization takes the distribution of SAR target images into account and softens the greedy search process which is highly prone to stick in local optima.
		
		\item We demonstrate that the SMGAA is generalizable across a variety of DNNs by comprehensive evaluations on eight cutting-edge DNN structures both in computer vision and SAR ATR applications, for example, 53.7$\%$, 77.1$\%$, and 87.2$\%$ average fooling rates can be respectively achieved by just 1, 2, and 3 adversarial scatterers. Compared with the currently studied attack algorithms, the SMGAA is shown to be robust to random noise, as well as Gaussian and median filters.
		
		\item We demonstrate the necessity and efficacy of the SMGAA in defending the malicious scatterers. The limitedly exploited SMGAA can bring robust accuracy improvements above 40$\%$ on average, distrust to the induced category, and consistent discriminative evidence to perturbed images, by evaluations on AConvNet. By contrast, current digital attack-based defense is shown to be almost ineffective against malicious scatterers.
		
	\end{enumerate}
	
	The remainder of this article is organized as follows. In Section \ref{relatedworks}, the background of adversarial attack and related works of this study are introduced. In Section \ref{imagformationofasc}, the ASCM and the image formation process based on it are elaborated. Our proposed adversarial scatterer generation algorithm is described in Section \ref{methodology}. Then, the experimental results and analysis are reported in Section \ref{experiment}, including the attack performance and application of defense, comparison to the $l_{p}$ attacks, and ablation study. In the end, Section \ref{conclusion} summarizes this article and designs the future work.

	\section{Background and Related Works}\label{relatedworks}
	In order to better comprehend the work of this article, we introduce the background linked to the adversarial attack for object recognition and the related works in SAR ATR.
	\subsection{Adversarial Attack in General Object Recognition}
	\subsubsection{Adversarial Attack}
	Adversarial attack aims at finding a visually imperceptible perturbation $\bm{\delta}$ with respect to an input image $\bm{x} \in \mathbb{R}^{h \times w}$ and its ground truth label $y^{\text{gt}} \in\mathbb{R}^{J}$, to mislead the classifier $F\!:\! \mathbb{R}^{h \times w}\! \rightarrow\! \mathbb{R}^{J}$ into outputting a wrong prediction $F(\bm{x}+\bm{\delta})\! \neq\! y^{\text{gt}}$ \cite{harnessing2015goodfellow,nguyen2015deep,szegedy2013intriguing}. This motivation can be described as the following problem:
	\begin{equation}
	\label{optimization}
	F(\underbrace{\bm{x}+\bm{\delta}}_{\bm{x}^{\text{adv}}}) \neq y^{\text{gt}}  \quad 	\text {s.t. } \mathcal{D}(\bm{x},\bm{x}+\bm{\delta}) \leq \epsilon
	\end{equation}
	where function $\mathcal{D}(\cdot)$ serve as a distance metric to guarantee that the perturbation $\bm{\delta}$ is imperceptible. There are several metrics have been studied, such as $l_{p}$-norm, geodesics in data manifold \cite{manifoldmetric}, Wassertain distance \cite{Wasserstein}, or perceptibility metrics \cite{functional}, \textit{etc}. In the remainder of this article, we focus on the most commonly studied $l_{p}$ attacks of which the perturbations are restricted in the $l_{p}$-norm ball with radius $\epsilon$, \textit{i.e.}, $\left\|\bm{\delta}\right\|_{p} \leq \epsilon$ and the $l_{p}$ norm $\left\|\cdot\right\|_{p}$ is defined as
	\begin{equation}
	\begin{aligned}
	\|\bm{\delta}\|_{p}  =\left(\sum_{\substack{0 \leq i<h \\ 0 \leq j<w}}\left|\delta_{i,j}\right|^{p}\right)^{\frac{1}{p}} &\stackrel{p=\infty}{\Longrightarrow} \max_{\substack{0 \leq i<h \\ 0 \leq j<w}}\left|\delta_{i j}\right| \\
	&\stackrel{p=0}{\Longrightarrow} \sum_{\substack{0 \leq i<h \\ 0 \leq j<w}} \mathbb{I} \left(\delta_{i j} \neq 0 \right), 
	\end{aligned}
	\end{equation}
	where $\mathbb{I}(\cdot)$ represents the indicator function. Generally $p$ is selected as 0 for sparsity, 2 for stealthiness, and $\infty$ for efficiency. When $p\!\neq\!0$, the classification loss can be efficiently magnified via normal gradient ascent process:
		\begin{equation}
		\boldsymbol{x}_{i+1}^{\text {adv}}=\boldsymbol{x}_{i}^{\text {adv }}+\epsilon_{i} \cdot \frac{\nabla_{x} \mathcal{L}\left(\boldsymbol{x}_{i}^{\text {adv }}, y^{\text {gt}}\right)}{\|\nabla_{x} \mathcal{L}\left(\boldsymbol{x}_{i}^{\text {adv}}, y^{\text {gt}}\right)\|_{p}}
		\end{equation}
		In the case of $l_{\infty}$ attacks such as FGSM, Basic Iterative Method (BIM) \cite{kurakin2016adversarial} and Projected Gradient Descent (PGD) \cite{madry2018towards}, the normalization can be replaced by the simple sign function. In addition to find perturbations with specific magnitude, some attacks pursue the minimal perturbations. For instance, C\&W \cite{carlini2017towards} proposed several customized loss functions to jointly optimize the $\mathcal{D}(\bm{x},\bm{x}+\bm{\delta})$ and classification loss. DeepFool \cite{moosavi2016deepfool} analytically derives the minimal perturbation that pull the sample across the decision hyperplane with the linear assumption of DNNs. The aforementioned attacks generate full-scale perturbations for the DNN inputs. $l_{0}$ attacks, by contrast, carefully perturb the input with limited pixels. There are currently sparse attacks based on variants of gradient-based methods integrating the $l_{0}$ constraint such as the PGD ($l_{0}$) \cite{croce2019sparse} and Jacobian-based Saliency Map Attack \cite{jsma} (JSMA), or search based methods like the Sparse-RS \cite{croce2020sparse}. Fig. \ref{fig1}-(a) exhibits the adversarial examples generated by different values of $p$ for a SAR target image. 
	
	Up to now, the SOTA DNNs are still extremely vulnerable to adversarial examples \cite{akhtar2018threat,serban2020adversarial}, highlighting the importance of adversarial robustness on security-critical tasks, such as autonomous driving and homeland security applications.
	There are many hypotheses tried to explain this non-robust behavior. Szegedy \textit{et al.} proposed that the adversarial examples represent the low-probability pockets in high-dimensional manifold \cite{szegedy2013intriguing}. It was also proposed that the existence of such examples is due to the strong linear nature of DNNs  \cite{harnessing2015goodfellow}. Tanay \textit{et al.} suggested that this odd behavior may be caused by the tilting decision boundary in high dimensional space \cite{tanay2016boundary}. 
	Although being a security threat, the existence of adversarial examples exhibits important value for enhancing the understanding \cite{ortiz2021optimism}, robustness, and trustworthiness of DNN-based high-dimensional classification \cite{serban2020adversarial}.
	
	\subsubsection{Adversarial Attack in Physical Setting}
	Many works in the field of optical imaging are devoted to revealing and describing real-world threats. The access to physically realizable perturbation mainly lies in three paths: 1) generating localized perceivable patch, \textit{e.g.}, pasting an adversarial sticker on the traffic sign\cite{liu2019perceptual,karmon2018lavan} or wearing a customized glass coated with adversarial pattern \cite{sharif2016accessorize}; 2) using infrared \cite{zhou2018invisible}, laser beam \cite{Duan_2021_CVPR} or exploiting the radiometric effect such as rolling shutter \cite{Sayles_2021_CVPR} to deploy perturbation that invisible to the human visual system; 3) constructing differentiable renderer \cite{liu2018beyond} to guide physical attack, which modifies the physical properties like the material, light, and geometry, \textit{etc}. Physical attacks in an optical imaging setting provide many insightful ideas, however, they cannot be directly utilized in SAR applications due to the different imaging mechanisms.

	\subsection{Adversarial Attack in SAR ATR}
	In contrast to the considerable studies in computer vision, the research on adversarial robustness of DNN-based SAR ATR is quite restricted. Typically, the study on the adversarial attack in SAR ATR has gone through two steps: 1) exploration and 2) optimization.
	Primeval research has focused on verifying the vulnerability of SAR models \cite{sarAAempirical2021,huangjnca,sarAAexperience2020} by using attack algorithms that borrowed from computer vision community. Works in this stage constructed comprehensive evaluations to reveal the adversarial vulnerability and also concluded some fundamental characteristics. For examples, Chen \emph{et al.} \cite{sarAAempirical2021} concluded that the adversarial examples transfer well across disparate SAR ATR models, and Li \emph{et al.} \cite{sarAAexperience2020} suggested that the vulnerability of models is proportional to their complexity. Additionally, it has demonstrated that the MSTAR targets have the intriguing trait of having misclassified classes of various models and attacks that are highly clustered. Afterwards, specific requirements in fooling SAR ATR motivate scholars to optimize the existing attack algorithms. For real-time perturbation generation, Du \emph{et al.} \cite{fastcw} accelerated the C$\&$W attack by introducing the generative models to learn the mapping from the original images to the adversarial examples. Peng \textit{et al.} proposed to enhance the attacks by manipulating the speckle noise pattern for a more non-cooperative condition \cite{sva}. However, these research concentrated on $l_{p}$-norm attack framework, which generates strictly pixel-wise perturbations and is less relevant to SAR imaging. Most related, the initial application of the parametric model in designing perturbation was of Dang \textit{et al.} published in 2021 \cite{9618140}, where the authors reported a 42$\%$ fooling rate against a support vector machine (SVM) based binary classification task.
	
	\section{Image Formation of the ASCM}\label{imagformationofasc}
	Based on high-frequency approximations \cite{keller1962geometrical}, the total backscatter can be decomposed into responses of individual scatterers. Developed from the geometric theory of diffraction (GTD) \cite{keller1962geometrical}, many parametric models were proposed for modeling the individual scatterers, including the Point Scattering Model (PSM) \cite{pointmodel}, Damped Exponential Model (DEM) \cite{dem}, and Attributed Scattering Center Model (ASCM) \cite{gerry1999parametric}, \textit{etc}. Among them, the ASCM characterizes both localized and distributed scattering mechanisms on frequency and aspect dependence as well as the physical attributes (such as the structure and size), provides a concise and physically relevant description for scatterers, and is compatible with most scattering models, and has shown great effectiveness in the state of the art (SOTA) SAR ATR techniques \cite{li2021multiscale,li2019sar,zhang2020fec}.
	
	In this section, the ASCM and its image formation process are introduced.
	
	\subsection{The ASCM}
	According to GTD \cite{keller1962geometrical,kouyoumjian1974uniform}, the summation of the responses from individual scatterers can well approximate the total backscattered field at the high-frequency region. Then, the ASCM describes the total scattered field as a function of frequency $f$ and aspect angle $\phi$ as \cite{gerry1999parametric}
	\begin{equation}
	E(f, \phi ; \bm{\Theta}_{N}) = \sum_{i=1}^{N} E_{i}\left(f, \phi ; \bm{\theta}_{i}\right)
	\end{equation}
	where $\bm{\Theta}_{N} \!=\! \left\{\bm{\theta}_{i} \mid \bm{\theta}_{i}= [A_{i}, x_{i}, y_{i}, \alpha_{i}, \gamma_{i}, L_{i}, \bar{\phi}_{i} ], 1 \leq i \leq N \right\}$ is the parameter set of $N$ individual scatterers and
	\begin{equation}
	\label{ASC}
	\begin{aligned}
	E_{i}&\left(f, \phi ; \bm{\theta}_{i}\right) \\
	&= A_{i} \cdot \left(j \frac{f}{f_{c}}\right)^{\alpha_{i}} \cdot \exp \left(-j \frac{4 \pi f}{c}\left(x_{i} \cos \phi+y_{i} \sin \phi\right)\right) \\
	&\cdot \operatorname{sinc}\left(\frac{2 \pi f}{c} l_{i} \sin \left(\phi-\bar{\phi}_{i}\right)\right) \cdot \exp \left(-2 \pi f \gamma_{i} \sin \phi\right).
	\end{aligned}
	\end{equation}
	Herein, $f_{c}$ is the center frequency of radar wave, $c$ is the velocity of light. For the $i$th scatterer, $A_{i}$ is amplitude, $x_{i}$ and $y_{i}$ are range and cross-range locations. $\alpha \in [-1,-0.5,0,0.5,1]$ models the frequency dependence.  For localized scattering mechanism that has localized returns in the SAR image, $L=\bar{\phi}=0$ and $\gamma$ describes the aspect dependence. For distributed scattering mechanism whose response span several image pixels, $\gamma=0$ and $[L ,\bar{\phi}]$ respectively model the length and orientation angle. Table \ref{scatterstype} lists the most common scattering structures in the real world, demonstrating the generalization ability of the ASCM. It can be seen that different types of geometric scattering structures can be distinguished by frequency dependence and length. For example, the dihedral degrades to trihedral when $L\!=\!0$. The parameter setting is important to ensure the physical feasibility, \textit{e.g.}, too large frequency response may not be achievable with existing materials or normal structures. Therefore, the parameter sensitivity will be further analyzed in the experiments.
	\begin{table}[tp] 
		\renewcommand{\arraystretch}{1.2}
		\centering
		\caption{Geometric scattering types differentiated by frequency and aspect dependence}
		\label{scatterstype}
		\begingroup
		\begin{tabular}{cccccc}
			\toprule[1.5pt]
			&\makecell{Geometric \\ scattering type} & $\alpha$ &$\gamma$  & $L$ & $\bar{\phi}$   \\
			\midrule
			\multirow{4}*{Localized} &  Trihedral  & 1 & \multirow{4}*{\textgreater 0} & 
			\multirow{4}*{0} & 
			\multirow{4}*{0} \\
			&  Top Hat (TH)  &  0.5   & &&  \\
			&  Sphere  &   0  & &&  \\
			&  Corner Diffraction (CD)  &  -1   &&& \\
			\hline
			\multirow{4}*{Distributed} &  Dihedral  &  1 & \multirow{4}*{0} & 
			\multirow{4}*{\textgreater 0} & \multirow{4}*{$\neq 0$}  \\
			&  Cylinder  &   0.5  &  & & \\
			&  Edge Broadside (EB)  &  0   & && \\
			&   Edge Diffraction (ED) &  -0.5   &&& \\
			\bottomrule[1.5pt]
		\end{tabular}%
		\endgroup
	\end{table}%
	\begin{figure}[b]
		\centering
		\subfigure[]{\includegraphics[width=5.8cm]{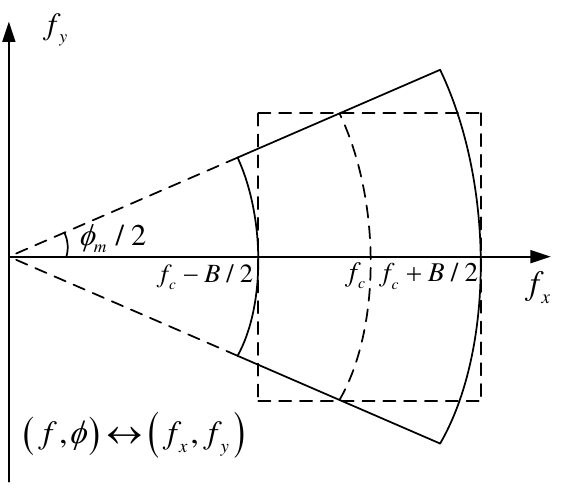}}
		\subfigure[]{\includegraphics[width=2.7cm]{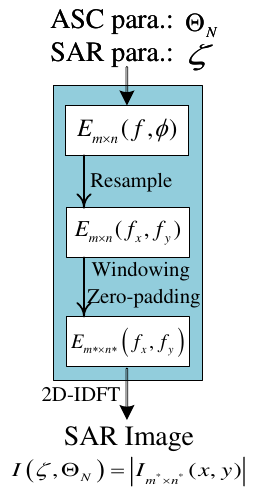}}
		\caption{(a) Resample the polar format data to the rectangular grid in the Cartesian plane.  (b) Image formation of the ASCM.}
		\label{imaging}
	\end{figure}
	
	\subsection{Image Formation}
	The response data calculated by Eq. \eqref{ASC} needs to be further coped to form the SAR image. 
	Fig. \ref{imaging} provides a general SAR image formation process of ASCM that was given in \cite{akyildiz1999scattering}. In frequency domain, the polar format data is uniformly sampled in frequency $f \in [f_{c}-B/2, f_{c}+B/2]$ and aspect angel $\phi \in [-\phi_{m}/2, \phi_{m}/2]$, where $B$ is the bandwidth of radar wave and $\phi_{m}$ is the aperture accumulation angle. Using $E_{m \times n}(f, \phi)$ to represent the total response in polar plane, which is formed by sampling $m$ and $n$ points in frequency and aspect angle sections, respectively.  First, $E_{m \times n}(f_{x}, f_{y})$ is obtained by uniformly resampling $E_{m \times n}(f, \phi)$ to the Cartesian plane where
	\begin{equation}
	\begin{aligned}
	f_{c}-B/2 \leq &f_{x}=f\cos(\phi) \leq f_{c}+B/2, \\
	-f_{c}\cdot\sin(\phi_{m}/2)\leq &f_{y}=f\sin(\phi) \leq f_{c}\cdot\sin(\phi_{m}/2)\\
	\end{aligned}
	\end{equation}
	To improve the quality of the resultant image, $E_{m \times n}(f_{x}, f_{y})$ will be multiplied by a window function $W(f_{x},f_{y})$ and then be zero-padded to $m^{*} \!\times\! n^{*}$ samples. Finally, the SAR image $I_{m^{*} \times n^{*}}(x, y)$ is obtained by 2D-IDFT. Through this process, the pixel spacing of range and cross-range are given by:
	\begin{equation}
	p_{x}=\frac{c}{2B} \cdot \eta_{x}, \quad p_{y}=\frac{1}{f_{c}} \cdot \frac{c}{4 \sin \left(\phi_{m} / 2\right)} \cdot \eta_{y},
	\end{equation}
	where $\eta_{x}=\frac{m-1}{m^{*}-1}$ and $\eta_{y}=\frac{n-1}{n^{*}-1}$ are the zero-padding effect.
	
	\subsection{Parameter Normalization} \label{normalizaiton}
	In Eq. \eqref{ASC}, $x$, $y$, and $L$ model the range and cross-range locations, and length of the scatterer in meters, and $\bar{\phi}$ measures the orientation angle in degrees. In addition, the aspect dependence factor $\gamma$ is smaller than other parameters by almost ten order of magnitude, which is not conducive to be updated. For the better numerical properties, and providing a pixel-level description of the scatterers' location and length, the model parameters are normalized using the following transformations \cite{akyildiz1999scattering,zab}
	\begin{equation}
	\begin{aligned}
	x_{p}=\frac{x}{p_{x}},  y_{p}=\frac{y}{p_{y}},  L_{p}=\frac{L}{p_{y}}, \gamma_{p}=\gamma\cdot 2\pi f_{c}, \bar{\phi}_{p}=\frac{\bar{\phi}}{\phi_{m}/2}.
	\end{aligned}
	\end{equation}
	By this way, $x_{p}$, $y_{p}$, are the pixel location and $L_{p}$ is pixel length, which is consistent with normal pixel index approach. The orientation index $\bar{\phi}_{p}$ is then lies into $[-1,1]$ and the normalized ASCM can be tidied as follows
	\begin{equation}
	\begin{aligned}
	E&\left(f_{x}, f_{y}; \bm{\theta}\right) \\
	&= A \cdot\left( \frac{j \sqrt{\left(f_{x}^{2}+f_{y}^{2}\right)}}{f_{c}}\right)^{\alpha} \cdot \exp \left(-\frac{f_{y}}{f_{c}}\gamma_{p}\right)\\
	&\cdot \operatorname{sinc}\left(\frac{\pi \sqrt{\left(f_{x}^{2}+f_{y}^{2}\right)}}{2 \sin \left(\phi_{m} / 2\right)f_{c}} L_{p} \eta_{y} \sin \left(\tan ^{-1}\left(f_{y} / f_{x}\right)-\bar{\phi}_{p} \phi_{m} / 2\right)\right)\\
	&\cdot \exp \left( \frac{-j4 \pi }{c}\left(p_{x} x_{p} f_{x}+p_{y} y_{p} f_{y}\right)\right).
	\end{aligned}
	\end{equation}
	
	To summarize at the end of this section, for a given parameter set $\bm{\Theta}_{N} \!=\!\left\{\bm{\theta}_{i} \mid \bm{\theta}_{i}= [A_{i}, x_{pi}, y_{pi}, \alpha_{i}, \gamma_{pi}, L_{pi}, \bar{\phi}_{pi}], 1 \leq i \leq N \right\}$ of the normalized ASCM, and the SAR imaging parameters $\bm{\zeta}=\left[f_{c}, B, \phi_{m}, m, n, m^{*}, n^{*}, W(\cdot) \right]$, the corresponding SAR image $\left|I_{m^{*} \times n^{*}}(x, y)\right| \rightarrow I(\bm{\zeta},\bm{\Theta}_{N})$ can be obtained using the imaging process illustrated in Fig. \ref{imaging}.

	\section{Crafting Adversarial Scatterers}\label{methodology}
	In this section, we detail the second part of the proposed SMGAA, including optimization objective of the normalized ASCM parameters and the customized strategies.
	
	\subsection{Objective}
	Recall the motivation of adversarial attack described in Eq. \eqref{optimization} and our motivation to generate adversarial scatterers. The perturbation $I(\bm{\zeta},\bm{\Theta}_{N})$ is controlled by the parameter set of normalized ASCM. Thus, the goal is to optimize the parameter set $\bm{\Theta}_{N}$ that makes the adversarial example $\bm{x}^{\text{adv}}\!=\!\bm{x}+I(\bm{\zeta},\bm{\Theta}_{N})$\footnote[2]{This image-level fusion is equivalent to signal-level superposition due to the linear nature of the discrete Fourier transform.} misclassified by the SAR target classifier at a very low cost, \textit{i.e.}, by adding a few adversarial scatterers. In addition to limiting the number of scatterers, a pair of vectors $[\bm{\theta}_{\text{max}},\bm{\theta}_{\text{min}}]$ is selected to restrict the parameter set in a proper range. Then, adversarial scatterers could be calculated by maximizing the classification loss:
	
	\begin{equation}\label{obj}
	\begin{aligned}
	\mathop{\arg\max}\limits_{\bm{\Theta}_{N}}  \;& \mathcal{L}(\bm{x}+I(\bm{\zeta},\bm{\Theta}_{N}),y^{\text{gt}}) \\ 
	\text { s.t. } &\bm{\theta}_{\text{min}}\leq \bm{\Theta}_{N} \leq \bm{\theta}_{\text{max}}\\
	& \bm{x}+I(\bm{\zeta},\bm{\Theta}_{N}) \in \mathbb{R}^{h \times w}\\
	& N \leq \epsilon 
	\end{aligned}
	\end{equation}
	where $\mathcal{L}(\cdot)$ calculates the loss of classifier $F$ with respect to the input $\bm{x}$ and its ground truth label $y^{\text{gt}}$. Here, the term $\bm{x}+I(\bm{\zeta},\bm{\Theta}_{N}) \in \mathbb{R}^{h \times w}$ guarantees that the adversarial example falls within the reasonable input interval for the model.
	
	\subsection{Method}
	Generally, approximate solutions of Eq. \eqref{obj} can be given by gradient-based methods such as BIM \cite{kurakin2016adversarial} and PGD \cite{madry2018towards} since the imaging function $I(\cdot)$ can be formulated differentiable. For a given scattering type, ASCM allows us to manipulate the scatterer parameters (attributes), such as amplitude, location, size and aspect angle. However, adjusting the images from these many modalities requires taking into account the impact of each parameter on the loss surface, which is a highly non-linear and non-convex task. Meanwhile, the optimization is prone to be stuck into local optima. To efficiently find adversarial scatterers, three simple, yet effective strategies are proposed to exploit the gradient. Algorithm \ref{alg1} elaborates the suggested procedure, which will be expanded upon in the following subsections.
	
	\begin{algorithm}[htbp]
		\caption{SMGAA}\label{alg1}
		\renewcommand{\algorithmicrequire}{\textbf{Input:}}
		\renewcommand{\algorithmicensure}{\textbf{Output:}}
		\begin{algorithmic}[1]
			\Require Classifier $F$; Image $\bm{x}$, corresponding category $y^{\text{gt}}$, and the target$\&$shadow mask $\bm{m}$; Imaging parameters $\bm{\zeta}$; Available range for ASC parameters $[\bm{\theta}_{\text{max}},\bm{\theta}_{\text{min}}]$; Maximum iteration $n^{\text{max}}$; Batch capacity $B$; Confidence threshold $v_{\text{th}}$; Normal distribution $\mathcal{N}(\bm{S}, \bm{\sigma})$; Learning rate $\lambda$; Number of scatterers $N$;
			\Ensure A parameter set $\bm{\Theta}_{N}$;
			\For{$j\!\gets\!1$ to $B$ and $i\!\gets\!1$ to $N$}  
			\State Initialize $\alpha_{i}^{j}$ and corresponding adjustable vector $\bm{\tau}^{j}_{i}$
			\State Initialize $x_{pi}^{j},y_{pi}^{j}$ with random coordinate pairs in $\bm{m}$
			\State Initialize other parameters with  $\operatorname{randu}[\bm{\theta}_{\text{min}},\bm{\theta}_{\text{max}}]$
			\EndFor
			\State $n \gets 1$
			\State $\bm{v} \gets C_{F}(\bm{x}+I(\bm{\zeta},\bm{\Theta}_{N}), y^{\text{gt}})$
			\State $\mathcal{L}\gets \mathcal{L}(\bm{x}+I(\bm{\zeta},\bm{\Theta}_{N}),y^{\text{gt}})$
			\While{$n<n^{\text{max}}$ or $\operatorname{min}(\bm{v})>v_{\text{th}}$}
			\State $n \gets n+1$
			\State $\Delta_{\bm{\Theta}}\gets\operatorname{randn}(\bm{S}^{i},\bm{\sigma})\cdot\operatorname{sign}(\nabla_{\bm{\Theta}}\mathcal{L})$
			\State $\bm{\Theta}_{*}=\operatorname{Clip}(\bm{\Theta}_{N}+ \bm{\tau} \cdot \Delta_{\bm{\Theta}},\bm{\theta}_{\text{min}},\bm{\theta}_{\text{max}})$
			\For{$i, j \in \{\operatorname{max}(I(\bm{\zeta},\bm{\Theta}_{*i}^{j})) > 1\}$}
			\State $A^{j}_{*i} \gets \frac{A^{j}_{*i}}{A^{j}_{*i}+\operatorname{randu}(1)}$
			\EndFor
			\State $\bm{x}^{\text{adv}} \gets \operatorname{Clip}(\bm{x} + I(\bm{\zeta},\bm{\Theta}_{*}), 0, 1)$
			\label{clip}
			\State $\mathcal{L}_{*}\gets \mathcal{L}(\bm{x}^{\text{adv}},y^{\text{gt}})$
			\For{$k\!\gets\!1$ to $B$}
			\If{$\mathcal{L}_{*}^{k} > \mathcal{L}^{k}$ or $\operatorname{randu}(1)>0.5$}
			\State $\bm{\Theta}_{N}^{k} \gets \bm{\Theta}_{*}^{k}$
			\If{$\mathcal{L}_{*}^{k} > \mathcal{L}^{k}$}
			\State $\bm{S}^{k}\gets \lambda\cdot\bm{S}^{k}+ (1-\lambda)\cdot \left| \Delta\bm{\Theta}^{k} \right|$
			\EndIf
			\EndIf
			\EndFor 
			\State $\mathcal{L}\gets \mathcal{L}_{*}$
			\State $\bm{v} \gets C_{F}(\bm{x}+I(\bm{\zeta},\bm{\Theta}_{N}), y^{\text{gt}})$
			\State idx $\gets\operatorname{argmin}(\bm{v})$
			\EndWhile
			\State \textbf{Return} $\bm{\Theta}_{N}^{\text{idx}}$
		\end{algorithmic}
	\end{algorithm}
	
	\subsubsection{Parameter Initialization}
	To assure the feasibility of the generated scatters, we choose the scattering types in Table \ref{scatterstype} to initialize the ASCM parameters, and the corresponding vectors are defined to guarantee the proper update. In specific, when initializing the candidate scatterers, the scattering types are determined with random types from Table \ref{scatterstype}. Meanwhile, a corresponding vector  $\bm{\tau}$ is generated to freeze the type-related parameters during the optimization such as the frequency dependence and length (for localized scatterers). The influence of the parameters is investigated in Section \ref{experiment}.
	It is known that the target and shadow parts of the SAR target image carry the majority of the structural information. In our context, the scatterers that are attached to the target and shadow regions are also more likely to deceive the DNN classifiers. Thus, the random coordinate pairs of the target and shadow mask $\bm{m}$ will be assigned to the initial scatterers. The other parameters are uniformly sampled in $[\bm{\theta}_{\text{min}}, \bm{\theta}_{\text{max}}]$ for better exploring the solution space. In this article, existing techniques are preferred to obtain the fine-grained mask $\bm{m}$.
	
	\subsubsection{Stepsize Adaption}
	The iterative gradient ascent marches by softened greedy search steps in which the fail update would also be randomly adopted to escape from the local optima:
	\begin{equation}
	\centering
	\begin{aligned}
	\Delta_{\bm{\Theta}}&=\bm{s}\cdot\operatorname{sign}(\nabla_{\bm{\Theta}}\mathcal{L})\\
	\bm{\Theta}_{N}^{*}&=\operatorname{Clip}(\bm{\Theta}_{N}+ \bm{\tau} \cdot \Delta_{\bm{\Theta}},\bm{\theta}_{\text{min}},\bm{\theta}_{\text{max}})\\
	\bm{\Theta}_{N} &= \begin{cases}\bm{\Theta}_{N}^{*}\quad\quad
	\text {if loss increases or $\operatorname{randu}(1)>0.5$} \\ 
	\bm{\Theta}_{N} \quad\quad \text{otherwise}\end{cases}
	\end{aligned}
	\end{equation}
	where $\mathcal{L}=\mathcal{L}(\bm{x} + I(\bm{\zeta},\bm{\Theta}_{N}),y^{\text{gt}})$ denotes the Cross-Entropy loss, $\operatorname{sign}(\cdot)$ determines in which direction the parameter set should be updated, and $\bm{s}$ is stepsize. Function $\operatorname{Clip}(\cdot, a, b)$ is applied to clip the input to range $[a, b]$. A Gaussian stepsize generator $\mathcal{N}(\bm{S}, \bm{\sigma})$ is designed to introduce randomness and capability of adaption during the optimization process. In each iteration, the stepsize $\bm{s}$ is generated by
	\begin{equation}
	\centering
	\bm{s} = \operatorname{randn}(\bm{S},\bm{\sigma}).
	\end{equation}
	The initial mean value $\bm{S}_{0}$ and standard deviation $\bm{\sigma}$ are pre-defined, and the mean value adapts across the iterations:
	\begin{equation}
	\begin{aligned}
	\bm{S}_{i+1} = \begin{cases}\lambda\cdot\bm{S}_{i}+ (1-\lambda)\cdot \left| \Delta\bm{\Theta} \right|
	& \text {if loss increases} \\ \bm{S}_{i} & \text {otherwise}\end{cases}
	\end{aligned}
	\end{equation}
	here factor $\lambda$ is the balance factor to relax potential overfitting of the adaption, and is set to 0.5 in the experiments. By utilizing this adaption strategy, a roughly defined stepsize can adapt diverse initialization and optimization processes, and the randomness brought by the generator can help escape from the local optimum.
	
	\subsubsection{Population-Style Generation Process}
\begin{figure}[tbp]
		\centering
		\includegraphics[width=0.95\hsize]{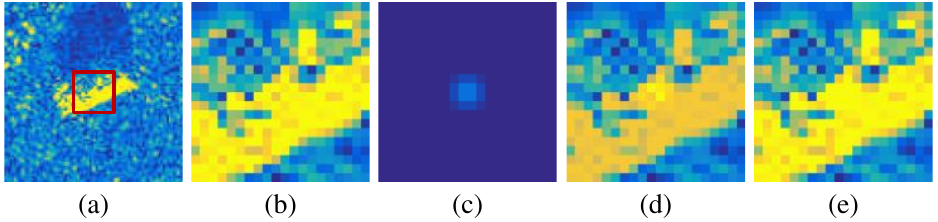}
		\caption{Illustration of the clip operation: (a) target image; (b) $20\!\times\!20$ central patch of (a) (normalized magnitude is 1); (c) a point scatterer response with normalized magnitude of 0.18; (d) fused image; (e) fused image after clip operation.}
		\label{clipfig}
	\end{figure}

	A population-style generation process is applied to accelerate the attack. Specifically, a batch of $B_{N}$ parameter sets, \textit{i.e.}, a $B_{N} \!\times \!N\!\times7$ tensor, would be initialized and optimized at once. Such a process reduces time consumption by increasing memory footprint. Notice that there has no information interaction between each individual since the generation process is considered mainly related to the initialization. During the updates, we control the amplitude of each adversarial scatterer lower than 1 (for normalized images), \textit{i.e.}, the response of the adversarial scatterers is not allowed stronger than the target. Practically, when the amplitude of each scatterer exceeds 1, the parameter $A$ would be divided by $A+\operatorname{randu}(1)$ to project the response back into the lower level. 
	Meanwhile, the adversarial examples will be clipped to $[0,1]$ before being fed into the model to satisfy the input range of DNNs. The clip operation mainly works when the adversarial scatterer is located at the strong response of the target, which is visualized in Fig. \ref{clipfig}. It can be observed that the scatterer response losses after clipping, leading to very limited additive perturbation to the original image. Nonetheless, this mild perturbation can be implemented by modifying the target, like by appending coatings that strengthen the scattering. Otherwise, the whole target will be suppressed by the local strong response across the maximum-normalization just as shown by Fig. \ref{clipfig}-(d), which is investigated in Section \ref{supperession}.
	Attack will stop when reaching the max iteration number $n^{\text{max}}$ and early stopping will be triggered when the confidence score of ground truth label $C_{F}(\bm{x}+I(\bm{\zeta},\bm{\Theta}_{N}), y^{\text{gt}})$ is lower than a threshold $v_{\text{th}}$.
	
	\section{Experiments}\label{experiment}
	Experiments on the publicly accessible dataset were conducted to evaluate the proposed method. Section \ref{setup} summarises the dataset and experimental setup.
	Section \ref{results} demonstrates the performance, analysis, and defense application of the SMGAA. 
	In Section \ref{compare}, the SMGAA is compared with the currently studied digital attacks in SAR community and the typical physical attack in the optical object recognition tasks.
	In Section \ref{ablation}, we investigate the efficacy of the proposed strategies and the parameter sensitivity of the generation algorithm. The discussion on the previously reported attack selectivity of MSTAR dataset is given in Section \ref{discussion}.
	\begin{table}[tb]
		\renewcommand{\arraystretch}{1.2}
		\centering
		\caption{Details of SOC subset of the MSTAR dataset}
		\label{table1}
		\begingroup
		\begin{tabular}{cccc}
			\toprule[1.5pt]
			Class & Serial Number&\makecell{Training Set\\ $(17^{\circ})$} & \makecell{Test Set\\		$(15^{\circ})$} 	\\
			\midrule
			2S1 & b01 &299 & 274\\
			BMP2 &9563 &233 & 195 \\
			BRDM2 & E71 &298 & 274 \\
			BTR70 & c71 &233 & 196 \\
			BTR60 & k10yt7532 &256 & 195 \\
			D7 & 92v13015 &299  &274 \\
			T72 & 132 &232 & 196 \\
			T62 & A51 &299 & 273 \\
			ZIL131 & E12 &299 & 274 \\
			ZSU234 & d08 &299 & 274 \\
			\midrule
			Total & &2747 & 2425\\
			\bottomrule[1.5pt]
		\end{tabular}%
		\endgroup
	\end{table}%
	\subsection{Dataset and Experimental Setup} \label{setup}
	\subsubsection{Dataset}
\begin{figure}[htbp]
		\centering
		\includegraphics[width=0.85\linewidth]{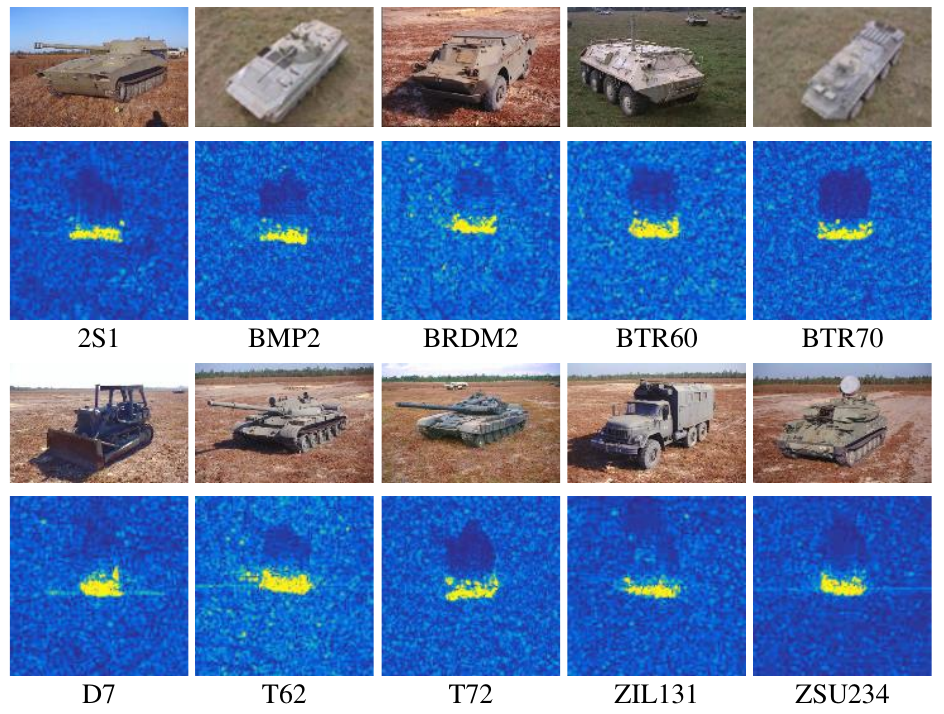}
		\caption{Examples of ten types ground vehicle targets in MSTAR dataset: 	(top) optical images and the corresponding (bottom) SAR images.}
		\label{mstar}
	\end{figure}

	The Moving and Stationary Target Acquisition and Recognition (MSTAR) program, with funding from the Defense Advanced Research Projects Agency (DARPA) and the Air Force Research Laboratory (AFRL), released a dataset \cite{mstar} for the public study of SAR ATR. The measured SAR data was collected using the Sandia National Laboratory SAR sensor platform with X-Band imaging ability in 1-foot resolution. The resulted dataset comprises SAR imagery with $360^{\circ}$ articulation with a $1^{\circ}$ spacing, an image size of $128^{2}$ pixels, and several depression angles. The MSTAR dataset contains ten types of ground vehicle targets (rocket launcher: 2S1; armored personnel carrier: BMP2, BRDM2, BTR70, BTR60; bulldozer: D7; tank: T62, T72; truck: ZIL131; air defense unit: ZSU234), the optical and SAR images of these targets are listed in Fig. \ref{mstar}. 
	
	There are four commonly studied partitions including one Standard Operate Condition (SOC) and three Extended Operate Conditions (EOC), respectively for the classification tasks of typical condition (10 targets), large depression variation (4 targets), configuration variants (5 variants of one target), and version variants (7 variants of 2 targets). The SOC subset was selected to carry out evaluations since this article mainly concentrates on category confusion. In the SOC subset, the images that had been collected at $17^{\circ}$ depression angle were used for training, and $15^{\circ}$ were used for testing. More specifics about the SOC subset are concluded in Table \ref{table1}.
	
	\begin{table}[htbp]
		\renewcommand{\arraystretch}{1.2}
		\centering	
		\caption{DNN models information}
		\begin{tabular}{ccccc}
			\toprule[1.5pt]
			Model & \makecell{Input Size} & $\#$ params. & \makecell{FLOPs \\($\times10^{9}$)} & \makecell{MSTAR \\Acc. ($\%$)}  
			\\ \midrule
			AlexNet \cite{alexnet2012} & $224^{2}$ &  58,299,082 & 1.06 & 96.3   \\
			VGG11 \cite{vgg2015sk} & $224^{2}$ & 128,814,154 & 2.02 & 98.1    \\
			ResNet50 \cite{resnet2016kh}  & $224^{2}$ & 23,522,250 & 4.04 & 96.4    \\
			DenseNet121 \cite{densenet2017huang} & $224^{2}$ & 6,957,706 & 2.80 & 97.8  \\
			MobileNetV2 \cite{mobilev2}& $224^{2}$  & 2,236,106  & 0.31 & 97.4  \\
			AConvNet \cite{aconvnet2016chen} & $88^{2}$  & 303,498  &  0.04 & 98.1   \\
			ShuffleNetV2 \cite{Ma_2018_ECCV} & $224^{2}$ & 1,263,422  & 0.14 & 96.9     \\
			SqueezeNet \cite{iandola2016squeezenet} & $224^{2}$ & 726,474 & 0.26  & 97.1    \\
			\bottomrule[1.5pt]
			\label{models}
		\end{tabular}
	\end{table}%
	\begin{table}[htbp]
		\renewcommand{\arraystretch}{1.2}
		\centering	
		\caption{Parameter setting of the proposed method}
		\begin{tabular}{cc}
			\toprule[1.5pt]
			Parameter & Value\\
			\midrule
			$n^{\text{max}}$ & 90 \\
			$B_{N}$ &100  \\
			$\bm{\theta}_{\text{min}}$ &   $[0, 0, 0, -1, 0, 0, -1]$ \\
			$\bm{\theta}_{\text{max}}$ &   $[10, 87, 87, 1, 2, 5, 1]$  \\
			$\bm{S}_{0}$ & $[0.05, 0.5, 0.5, 0, 0.01, 0.025, 0.01]$  \\
			$\bm{\sigma}$ & $(\bm{\theta}_{\text{max}}-\bm{\theta}_{\text{min}})/200$  \\
			$\lambda$ & 0.5\\
			$v_{\text{th}}$ & 0.1\\
			$\bm{\zeta}$ & [9.6, 0.59, 0.051, 85, 85, 128, 128, (Taylor, -35dB)] \\
			\bottomrule[1.5pt]
			\label{paras}
		\end{tabular}
	\end{table}%
	
	\subsubsection{DNN Structures}
	In the experiments, eight typical DNN structures were involved to access the comprehensive results. Among them, AlexNet\footnote[3]{\url{https://github.com/pytorch/vision/tree/master/torchvision/models}\label{torchvision}} \cite{alexnet2012}, VGG11\textsuperscript{\ref {torchvision}} \cite{vgg2015sk}, ResNet50\textsuperscript{\ref {torchvision}} \cite{resnet2016kh}, and DenseNet121\textsuperscript{\ref{torchvision}} \cite{densenet2017huang} are typical feature extraction backbones and have achieved eye-catching performance in various domains. In addition, MobileNetV2\textsuperscript{\ref {torchvision}} \cite{mobilev2}, AConvNet\footnote[4]{We build the AConvNet according to original paper with Pytorch framework. The authors' released Caffe code can be found at \url{https://github.com/fudanxu/MSTAR-AConvNet}} \cite{aconvnet2016chen},  ShuffleNetV2\textsuperscript{\ref {torchvision}} \cite{Ma_2018_ECCV}, and SqueezeNet\textsuperscript{\ref {torchvision}} \cite{iandola2016squeezenet} are lightweight designs of DNN models that reduce the number of parameters, model size, and the requirement for computing resources. The lightweight models are more appropriate for the circumstances which the computing resource is limited, such as the edge-device and on-board SAR image processing task.
	
	\subsubsection{Data Processing and Implementation Details}
	The proposed method and the eight involved DNN models were implemented using the Python (v3.6) and Pytorch deep learning framework (v1.10.1). All the experiments were supported by an NVIDIA DGX-1 server, which is powered by a dual 20-Core Intel Xeon E5-2698 v4 CPU and equipped with eight Tesla-V100 GPUs. In addition, all the experiments were accelerated by CUDA Toolkit (v10.2).
	
	The DNN models were trained following the the well-known preprocessing of MSTAR data \cite{aconvnet2016chen}. Specifically, all the target slices were normalized to $[0, 1]$ to accelerate the convergence of the loss function. The single-channel data was fed into the models as gray-scale images. The random $88^{2}$ patches of training images were used for training as data augmentation, and the central patches of test images were for testing the models' accuracy. The Cross-Entropy loss function and the Stochastic Gradient Descent (SGD) optimizer were employed to train all the studied models, with more details reported in Table \ref{models}. The publicly accessible MSTAR segmentation annotation SARBake\footnote[5]{SARBake is available at \url{https://data.mendeley.com/datasets/jxhsg8tj7g/3}} \cite{malmgren2015convolutional} was utilized as coordinates pool $\bm{m}$ to initialize the scatterers' location.
	\begin{figure}[htbp]
		\centering
		\includegraphics[width=0.85\hsize]{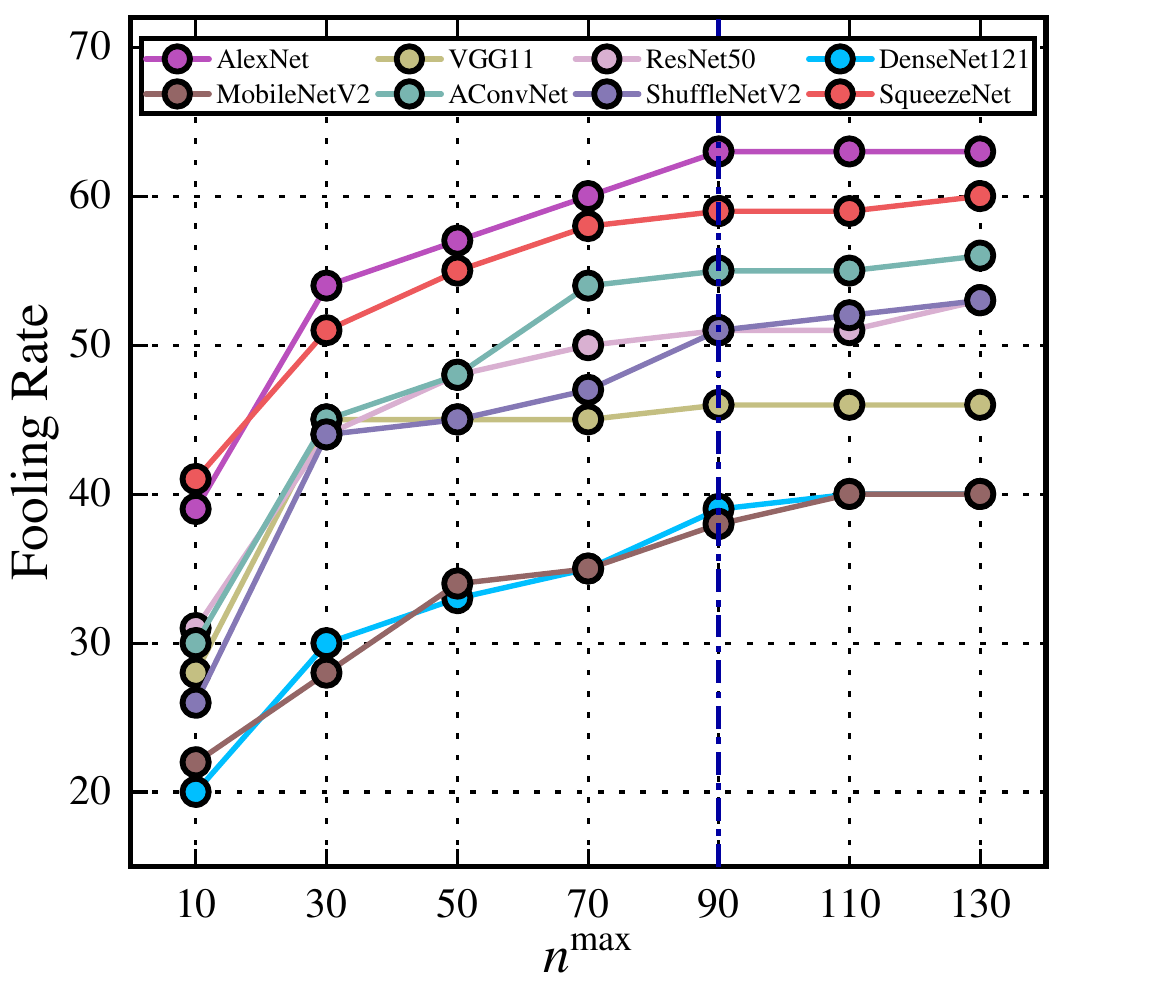}
		\caption{Fooling rate as a function of the maximum iteration $n^{\text{max}}$, which was obtained on 100 images with $B_{N}=50$. When increasing the maximum iterations, the fooling rates of SMGAA-1 against eight models increase and become level. According to the curves, $n^{\text{max}}=90$ was selected to perform following evaluations.}
		\label{parameterselection}
	\end{figure}
	
	The parameter setting of the generation algorithm is listed in Tab. \ref{paras}. Among them, $n^{\text{max}}$ was set to 90 according to a simple search that described in Fig. \ref{parameterselection}. Generally, the more likely the entire optimization process will succeed when batch capacity $B_{N}$ is larger. And it becomes an exhaustive search when $B_{N}\!\rightarrow\!\infty$. To seek a balance between efficacy and computational burden, $B_{N}$ was set to 100 to obtain candidate scatterers of good quality. Notice that $\bm{S}_{0}$ and $\bm{\sigma}$ 
	were just roughly set as $(\bm{\theta}_{\text{max}}-\bm{\theta}_{\text{min}})/200$. The analyses on the effect of ASCM parameters to the adversarial scatterers and the generation algorithm are provided respectively in Section \ref{tolerance} and \ref{ablation}. The SAR parameter set $\bm{\zeta}$ was calculated according to the file of the MSTAR dataset.
	\begin{table*}[tbp] 
		\renewcommand{\arraystretch}{1.2}
		\centering
		\caption{Fooling rates ($\%$) against eight DNN models achieved by SMGAA-1, -2, and -3 respectively}
		\label{untargeted}
		\begin{tabular}{cccccccccc}
			\toprule[1.5pt] 
			& AlexNet  & VGG11  & ResNet50  & DenseNet121& MobileNetV2  & AConvNet   & ShuffleNetV2& SqueezeNet & \textbf{Average}\\
			\midrule
			SMGAA-1 & 67.3 & 49.5 & 50.2 & 43.8 & 39.5 & 60.4 & 59.2 & 59.6 & 53.7 \\
			SMGAA-2 & 87.4 & 78.8 & 73.2 & 69.0 & 62.8 & 82.4 & 79.4 & 83.9 & 77.1 \\
			SMGAA-3 & 93.0 & 89.8 & 85.3 & 81.7 & 74.8 & 91.9 & 88.8 & 92.3 & 87.2\\
			\bottomrule[1.5pt]
		\end{tabular}%
	\end{table*}%
	
	\begin{figure*}[tbp]
		\centering
		\subfigure[SMGAA-1]{\includegraphics[width=1\textwidth]{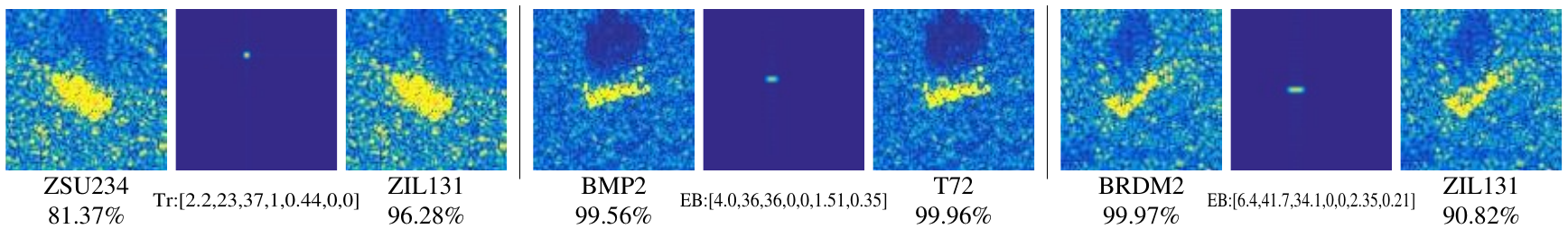}}\\
		\subfigure[SMGAA-2]{\includegraphics[width=1\textwidth]{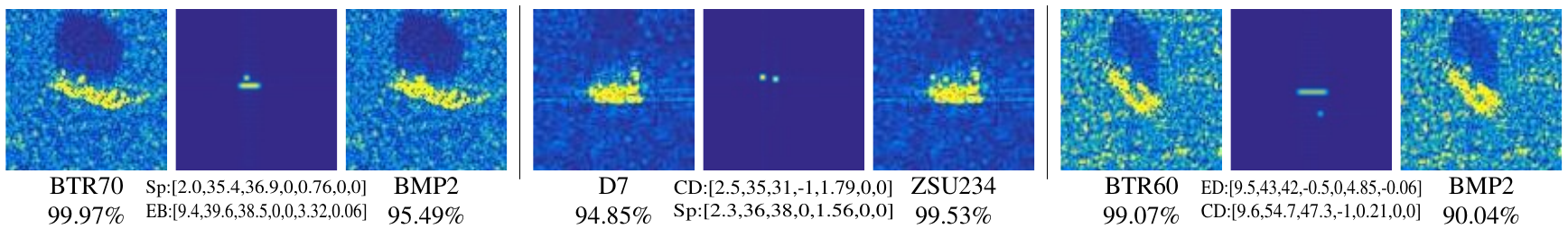}}\\
		\subfigure[SMGAA-3]{\includegraphics[width=1\textwidth]{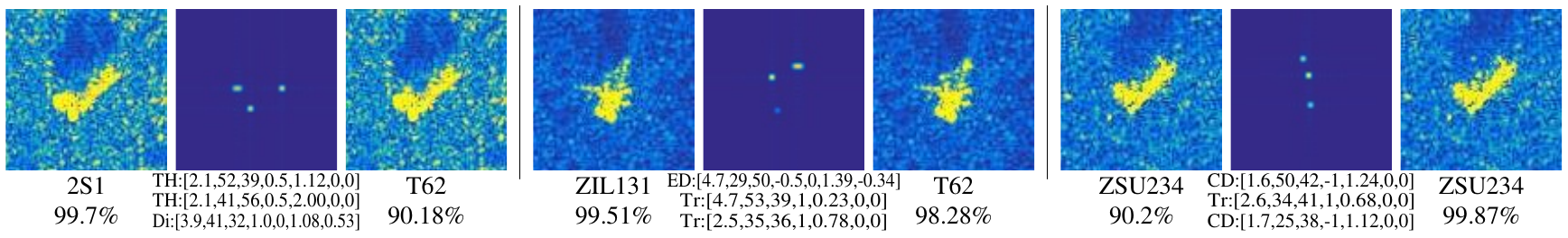}}\\
		\caption{Adversarial scatterers generated for AConvNet by SMGAA. Each of subplot groups contains the original image, adversarial scatterers, and adversarial example (from left to right). The prediction and corresponding confidence score are captioned under every target images. The scattering types and normalized ASCM parameter set corresponding to the adversarial scatterers are also elaborated.}
		\label{arc}
	\end{figure*}
	\subsubsection{Evaluation Measurements}
	The attack performance was measured using the fooling rate, which is defined as
	\begin{equation}
	\text {Fooling rate}=\frac{\sum_{i}^{N_{\text{total}}} \mathbb{I} \left( F(\bm{x}^{\text{adv}}_{i}) \neq y^{\text{gt}}_{i}\right) }{N_{\text{total}}}
	\end{equation}
	where $N_{\text{total}}$ is the capacity of the evaluated images in the experiments. The fooling rate denotes the ratio of misclassification after being attacked and the value is between 0 and 1. Fundamentally, with fair constraints and interference, the larger fooling rate indicates the stronger attack algorithm. 
	
	\subsection{Results} \label{results}
	A 1000-image subset was equally sampled from the test set to ensure the misclassification is indeed caused by attacks and to eliminate the effect of category imbalance, which contains a hundred images for each category and is correctly classified by all the DNNs. All the following results were obtained on this subset. For abbreviation, the SMGAA with $N$ adversarial scatterers is denoted by SMGAA-$N$ in the rest of this article.
	
	\subsubsection{Quantitative Results}
	We report the fooling rates achieved by SMGAA-$N$ attacks against the eight evaluated models in Table \ref{untargeted}, from which the following summaries can be obtained. 
	Firstly, the SMGAA is capable of achieving considerable fooling rates against both the studied general and lightweight DNN structures that even a single scatterer can achieve an average fooling rate of 53.7$\%$. Secondly, it shows that the more available scatterers to deploy the higher fooling rates can be achieved, illustrating the objective \eqref{obj} is effective in fooling the DNN-based SAR ATR. 
	
\begin{figure}[tbp]
		\centering
		\includegraphics[height=0.8\hsize]{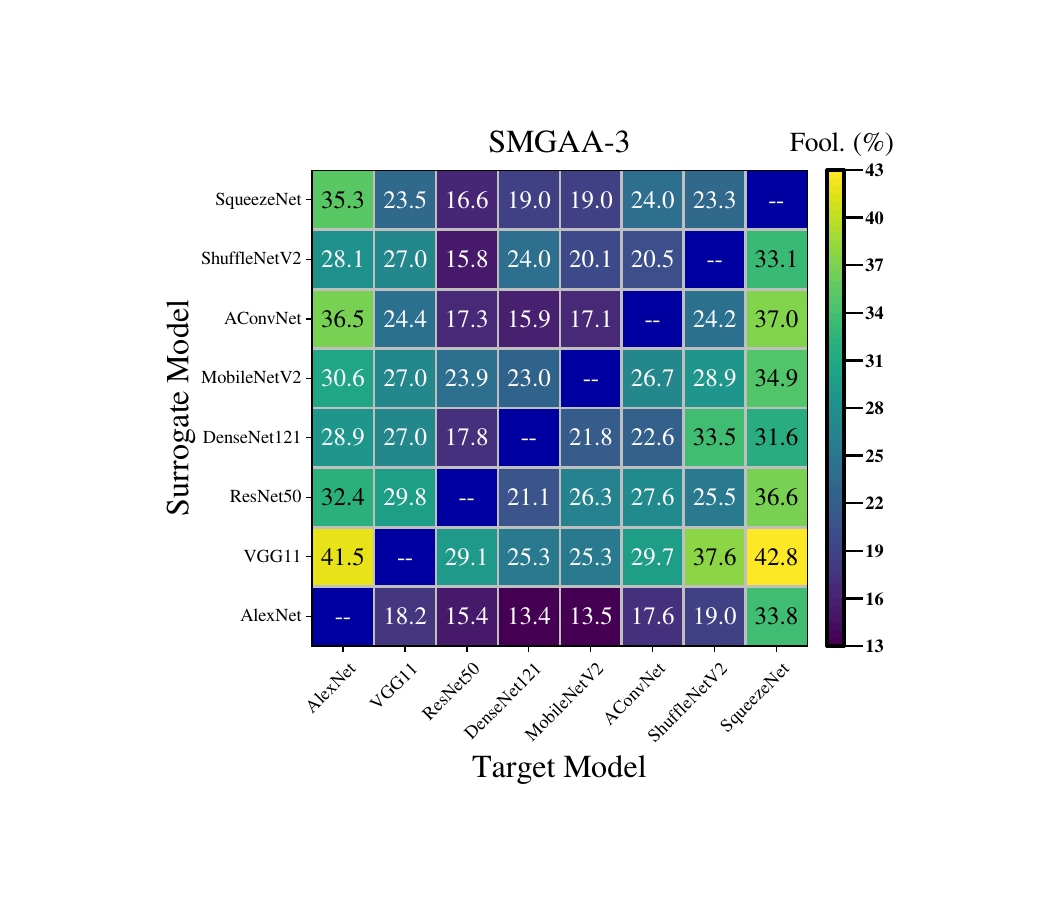}
		\caption{Transfer attack fooling rates of SMGAA-3. The \textit{Surrogate Model} refers to the one used to craft adversarial examples, and the \textit{Target Model} denotes the unknown target classifier.}
		\label{cf}
	\end{figure}

	The transfer attack was performed to verify the cross-model transferability of the SMGAA. Fig. \ref{cf} depicts the transfer fooling rates achieved by SMGAA-3 attack of which were obtained by attacking the black-box \textit{target model} using the adversarial examples generated by the \textit{surrogate model}. What stands out in the matrix is that the SqueezeNet is fooled by black-box SMGAA-3 with 35.7$\%$ fooling rate on average, and the adversaries crafted on VGG11 achieve an average fooling rate of 33.0$\%$ against other seven black-box models. It indicates that there is an adversarial threat present even in a highly non-cooperative setting and even with the simple addition of a few extra scatterers.
	
	\subsubsection{Visualizations}
	Some results of the SMGAA are visualized in Fig. \ref{arc}. In each subgroup, the original image, adversarial scatterer, and perturbed image are arranged from left to right, and the corresponding prediction and confidence score given by AConvNet, as well as the geometric scattering types, normalized ASCM parameters of the scatterers, are listed below the images. It can be observed that our attack framework is capable of achieving satisfactory visual stealthiness and powerful deception, that is, can fool the DNN classifier with a high confidence score (over 90$\%$, depending on the value of $v_{\text{th}}$). Moreover, the adversarial scatterers carry explicit physical attributes compared with the digital perturbations shown in Fig. \ref{fig1}. 
	
	\subsubsection{Parameter Sensitivity}\label{tolerance}
	\begin{figure}[tbp]
		\centering
		\includegraphics[width=1\hsize]{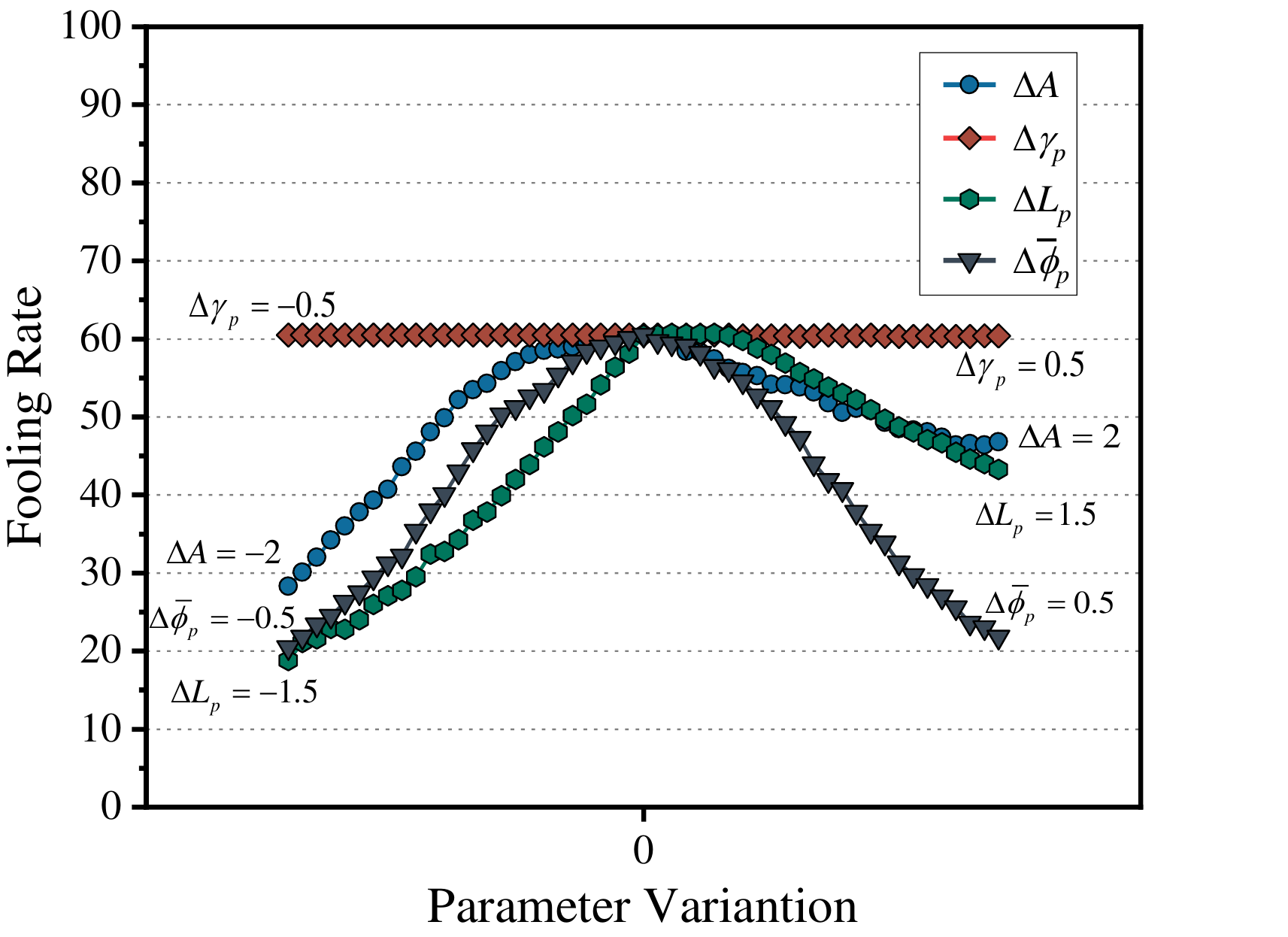}
		\caption{Fooling rate of SMGAA-1 against AConvNet as a function of the variation of $A$, $\gamma_{p}$, $l_{p}$, and $\bar{\phi}_{p}$.}
		\label{parasrobustness}
	\end{figure}
	\begin{figure}[tbp]
		\centering
		\includegraphics[width=0.9\hsize]{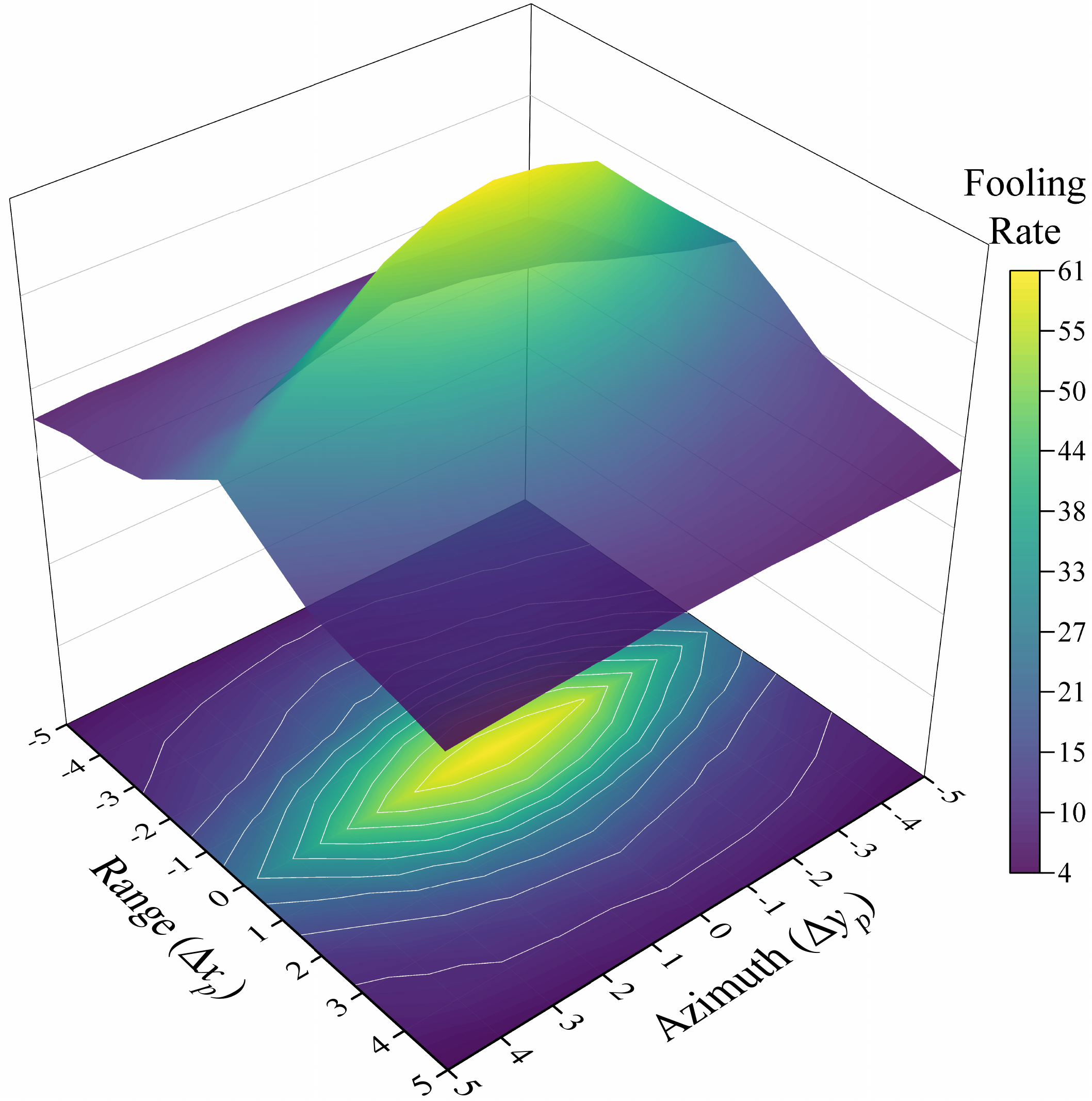}
		\caption{Fooling rate of SMGAA-1 against AConvNet as a function of the variation of $x_{p}$ and $y_{p}$.}
		\label{locationrobustness}
	\end{figure}
	To verify the parameter sensitivity of the scatterers, we modified their parameters and then re-tested the fooling rate. Fig. \ref{parasrobustness} depicts the sensitivities to $A$, $\gamma_{p}$, $L_{p}$, and $\bar{\phi}_{p}$, of which results were obtained on the resulted scatterers of SMGAA-1 for AConvNet. The effect of the location variation on both range and azimuth directions is shown in Fig. \ref{locationrobustness}. Several observations can be summarized. Firstly, the attack performance is not affected by the frequency\footnote[6]{When $\alpha$ fluctuates to each of the five possible values, the fooling rate remains constant.} and aspect dependencies. The reason is that $\alpha$ and $\gamma_{p}$ have too little effect on the resulted image. Secondly, the fooling rate decreases when the amplitude turns too larger or smaller, indicating that a certain intensity of scatterer is required to perform a successful attack. Thirdly, as the length increases, the fooling rate remains level and eventually decreases, since increasing length will disperse the total energy and lead to a lower intensity. Finally, the adversarial scatterers are more sensitive to excursion along range direction than along the azimuth direction. The reason is that the distributed scatterer has an extension in the azimuth direction and is more robust to small displacements.
	
	Overall, the results show that the adversarial scatterers are mainly affected by their intensity, location, length, and aspect angle. The attack performance is stable within a relatively narrow range of parameter fluctuation. It demonstrates the high practical relevance of SMGAA in designing physical adversarial scatterers, that is, attackers do not need to carefully engineer material properties to meet frequency and angle response requirements, while also having certain tolerance for other attributes.
	
	\subsubsection{Application of Defense}
\begin{figure}[tbp]
		\centering
		\includegraphics[width=0.85\hsize]{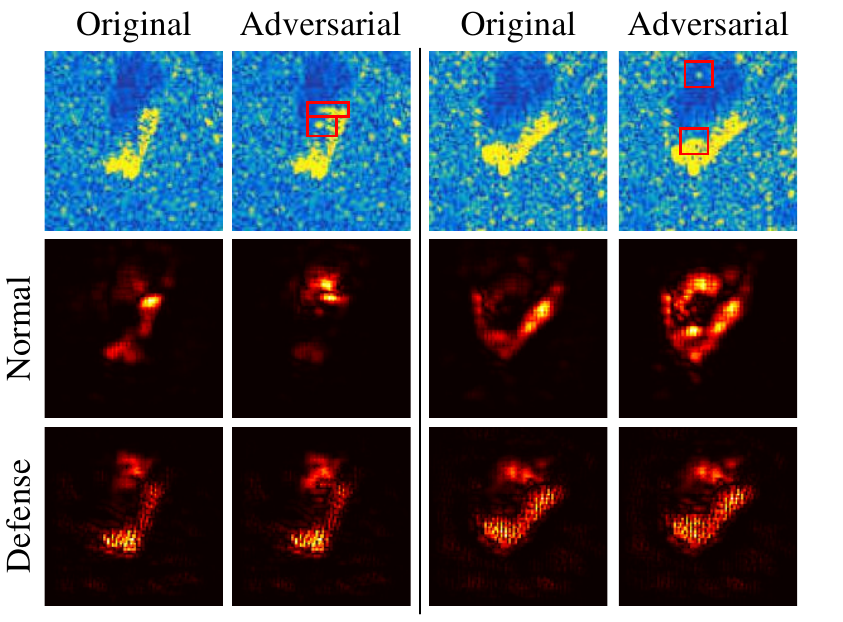}
		\caption{In both of the subplot groups, the pictures in the first column are the clean input images in turn, the Guided-Backpropagation \cite{springenberg2014striving} results of the input images extracted by the normally trained AConvNet, and the results extracted by the adversarial trained one. The illustrated adversarial examples were generated on VGG11 other than both the normally and adversarially trained AConvNets.}
		\label{robust}
	\end{figure}

	\begin{table}[tbp] 
		\renewcommand{\arraystretch}{1.2}
		\centering
		\caption{Comparison of AConvNet with and without defense. $\epsilon=4/255$ is set for PGD and other parameter setting can be found in Table \ref{comparesetting2}}
		\label{defense1}
		\begingroup
		\begin{threeparttable}
			\begin{tabular}{cccc}
				\toprule[1.5pt]
				& \multicolumn{3}{c}{Accuracy}    \\ 
				\cmidrule(r){2-4} 
				Attacks & Normal & \makecell{Defense (SMGAA)} & \makecell{Defense (PGD)}\\
				\midrule
				Clean & \textbf{98.1} & 97.4$\downarrow$ & 96.5$\downarrow$ \\
				SMGAA-1 &39.6& \textbf{80.8}$\uparrow$ & 39.3$\downarrow$  \\
				SMGAA-2 &17.6 & \textbf{62.5}$\uparrow$& 19.3$\uparrow$    \\
				SMGAA-3 &8.1 & \textbf{46.3}$\uparrow$&10.9$\uparrow$  \\
				PGD & 41.9 &13.4$\downarrow$  &  \textbf{86.5}$\uparrow$  \\
				\midrule
				&\multicolumn{3}{c}{Confidence on the adversarial category}   \\ 
				\cmidrule(r){2-4} 
				Attacks & Normal & \makecell{Defense (SMGAA)} & \makecell{Defense (PGD)}\\
				\midrule
				SMGAA-1  & 86.5 & \textbf{71.6}$\downarrow$ &  77.9$\downarrow$   \\
				SMGAA-2  & 90.4& \textbf{77.2}$\downarrow$ &82.9$\downarrow$ \\
				SMGAA-3  & 91.0 & \textbf{80.7}$\downarrow$ & 85.3$\downarrow$ \\
				PGD & 89.6 & 87.2$\downarrow$ & \textbf{66.6}$\downarrow$ \\
				\bottomrule[1.5pt]
			\end{tabular}%
			\begin{tablenotes}
				\item Note: The highest accuracies against the adversarial examples and the lowest confidences on the adversarial category are highlighted in \textbf{bold}.
			\end{tablenotes}
		\end{threeparttable}
		\endgroup
	\end{table}%
	It is our intention and the most direct application of the proposed method, that to defend against the randomly destructive or intentionally malicious scatterers. With the adversarial training framework \cite{inkawhich2020advanced,madry2018towards}, our proposal can be utilized as the worst-case scatterer generator to facilitate the min-max optimization of robust model training. Particularly, every training SAR target image will be fed into the training model with the adversarial scatterers. The general adversarial training procedure is expensive, thus, we set a efficient SMGAA-3 attack as the surrogate generator of whose batch capacity and iteration was set to 1 and 20. The comparisons between the normally and adversarially trained AConvNets are presented in Table \ref{defense1} and Fig. \ref{robust}.
	
	{In Table \ref{defense1}, we compare the SMGAA with the most famous adversarial training procedure with PGD attack. It reports the accuracy against SMGAA-$N$ and PGD attacks, and the confidence score on the successful adversarial examples. We can observe from the table that the adversarially trained model by SMGAA exhibits considerable robust accuracy improvements and distrust against SMGAA-$N$ attacks even just trained with limitedly exploited SMGAA-3 attack. We can also observe that the PGD-trained model is not capable of defending the adversarial scatterers, illustrating the necessity of the proposed method for defense purposes. Fig. \ref{robust} further visualizes the attention maps (extracted by Guided-Backpropagation \cite{springenberg2014striving}), contrasting the discriminative evidence of the normally and adversarially trained models. As can be seen the robust model gives consistent attention to the adversarial examples, while the non-robust one is significantly confused. 
	
	To summarise, our proposal is shown to be promising and effective for defense purposes. In addition to a compatible defense efficacy against more expensive attacks than the surrogate attack, the robust model also shows well generalizability to defend against attacks with different numbers of scatterers.
	
	\subsection{Comparative Experiments} \label{compare}
	\begin{table*}[htbp]
		\renewcommand{\arraystretch}{1.2}
		\centering
		\caption{Fooling rates and average transfer rates ($\%$) against eight DNN models achieved by various attack algorithms. The comparative algorithms include 1) $l_{\infty}$: FGSM \cite{harnessing2015goodfellow}, BIM \cite{kurakin2016adversarial}, and PGD \cite{madry2018towards}; 2) $l_{0}$: JSMA \cite{jsma}, Sparse-RS \cite{croce2020sparse}, and PGD ($l_{0}$)  \cite{croce2019sparse}; 3) $l_{2}$: SVA ($l_{2}$) \cite{sva}, DeepFool \cite{moosavi2016deepfool}, and C$\&$W \cite{carlini2017towards}; 4) physical attack: LAVAN \cite{karmon2018lavan}. The $l_{\infty}$ and $l_{2}$ constraint for three conditions were set to the mean perturbation caused by SMGAA-$N$. For $l_{0}$ attacks, the maximum number of pixels perturbed was set to the number of pixels that contain more than 99$\%$ of the adversarial scatterers' energy. The physical attack was appointed no constraint}
		\begin{threeparttable}
			\begin{tabular}{ccccccccccc}
				\toprule[1.5pt]
				\multicolumn{11}{c}{Fooling Rate}\\ 
				& Attacks 	& AlexNet  & VGG11  & ResNet50  & DenseNet121& MobileNetV2 & AConvNet   & ShuffleNetV2& SqueezeNet & \textbf{Average} \\
				\midrule
				\textbf{Proposed} & SMGAA-1 & 67.3 & 49.5 & 50.2 & 43.8 & 39.5 & 60.4 & 59.2 & 59.6 & 53.7 \\ \cline{2-11}
				\multirow{3}*{$l_{\infty}$} & FGSM \cite{harnessing2015goodfellow} & 24.1 & 5.2  & 10.1 & 59.2 & 78.9 & 2.2 & 73.4 & 4.8  & 32.2 \\
				& BIM \cite{kurakin2016adversarial} & 26.9 & 6.0 & 10.2 &79.9  & 99.6 & 2.3 & 96.5  & 5.1 & 40.8 \\ 
				& PGD \cite{madry2018towards} & 28.0 & 6.1 & 10.6 & 87.8 & \underline{\textbf{100.0}} & 2.3 & \underline{99.2} & 5.2 & 42.4\\ \cline{2-11}
				\multirow{3}*{$l_{0}$} & JSMA \cite{jsma} & - & - & - & - & - & 53.3 & -  & - & 53.3 \\
				& Sparse-RS \cite{croce2020sparse}   & 74.9 & 20.1 & 24.9 & \underline{97.7} & 79.0 & 17.6 & 94.0 &39.0 & 55.9 \\
				&PGD ($l_{0}$)  \cite{croce2019sparse} & \underline{90.1} & 21.0 & 30.4 & 92.7 & 76.8 & 40.0 & 96.6 & 77.1& 65.6 \\	
				\cline{2-11}
				$l_{2}$ & SVA ($l_{2}$) \cite{sva} & 85.0 & \underline{69.3} & \underline{78.2} & 92.9 & 78.6 & \underline{80.2} & 76.5 & \underline{79.6} & \underline{80.0}  \\ \midrule
				\textbf{Proposed} & SMGAA-2  & 87.4 & \underline{78.8} & 73.2 & 69.0 & 62.8 & 82.4 & 79.4 & 83.9 & 77.1 \\ \cline{2-11}
				\multirow{3}*{$l_{\infty}$} & FGSM \cite{harnessing2015goodfellow} & 40.6 & 10.0 & 17.9 & 77.3 & 85.3 & 3.6 & 83.0 & 7.9 & 40.7 \\
				& BIM \cite{kurakin2016adversarial}& 47.7 & 11.3 & 19.2 & 95.0 & \underline{\textbf{100.0}} & 3.7 & 99.8 & 8.4 & 48.1 \\
				& PGD \cite{madry2018towards} & 51.0 & 11.6 & 19.5 & \underline{97.9} & \underline{\textbf{100.0}} & 3.7 & 99.9 & 8.9 & 49.1 \\ \cline{2-11}
				\multirow{3}*{$l_{0}$} & JSMA \cite{jsma} & -  & - & - & - & - & 66.9 & - & - & 66.9\\
				& Sparse-RS \cite{croce2020sparse} & 80.3 & 28.4 & 33.6 & 95.5 & 85.8 & 25.5 & 97.9 & 45.1 & 61.5 \\
				&PGD ($l_{0}$) \cite{croce2019sparse}  & \underline{93.5} & 36.4 &44.8  &92.0  & 88.5 & 67.3 & \underline{99.2} & \underline{88.2} & 76.2 \\ \cline{2-11}
				$l_{2}$ & SVA ($l_{2}$) \cite{sva} & 87.7 & 75.0 & \underline{81.3} & 93.9 & 83.2 & \underline{84.9} & 79.5 & 82.8 & \underline{83.5} \\ \midrule 
				\textbf{Proposed} & SMGAA-3  & 93.0 & \underline{89.8} & \underline{85.3} & 81.7 & 74.8 & \underline{91.9} & 88.8 & \underline{92.3} & \underline{87.2} \\ \cline{2-11}
				\multirow{3}*{$l_{\infty}$} & FGSM \cite{harnessing2015goodfellow} & 54.3 & 13.6 & 25.4 & 85.7 & 89.1 & 4.9 & 86.4 & 11.0 & 46.3\\
				& BIM \cite{kurakin2016adversarial} & 62.3 & 16.3 & 27.7 & 98.3 & \underline{\textbf{100.0}} & 5.1 & 99.9 & 12.4 & 52.8\\
				& PGD \cite{madry2018towards} & 66.1 & 17.2 & 29.0 & \underline{99.6} & \underline{\textbf{100.0}} & 5.3 & \underline{\textbf{100.0}} & 12.9 & 53.8 \\ \cline{2-11}
				\multirow{3}*{$l_{0}$} & JSMA \cite{jsma} &-  & - & - & - & - & 71.7 & - & - & 71.7 \\
				& Sparse-RS \cite{croce2020sparse} & 81.8  & 32.6 & 35.9 & 94.6 & 87.8 & 29.7 & 97.7 & 47.2 & 63.4  \\ 
				&PGD ($l_{0}$) \cite{croce2019sparse}  & \underline{94.2} & 44.7 & 53.0 & 91.8 & 90.7 & 77.6 & \underline{\textbf{100.0}} & 91.2 & 80.4 \\  \cline{2-11}
				$l_{2}$ & SVA ($l_{2}$) \cite{sva} & 88.7& 78.1 & 82.9 & 95.2 & 85.2 & 87.5 &  82.2& 85.0 & 85.6 \\ \midrule
				\multirow{2}*{minimal $l_{2}$} & DeepFool \cite{moosavi2016deepfool} & 99.7 & 98.3 & 90.5 & 99.6 & 99.5 & 92.4 & 99.4 & 89.4 & 96.1 \\
				& C$\&$W \cite{carlini2017towards} & \textbf{100.0} & \textbf{99.1} & \textbf{99.6} & \textbf{100.0} & \textbf{100.0}  & \textbf{100.0} & \textbf{100.0} & \textbf{100.0} & \textbf{99.8}\\
				\midrule
				Physical & LAVAN \cite{karmon2018lavan} & 49.1 & 2.7 & 35.4 & 93.8 & 34.0 & 2.9 & 85.3 & 7.8 & 38.9 \\ 
				\midrule
				\multicolumn{11}{c}{Average Transfer Rate}\\
				\textbf{Proposed} & \multicolumn{3}{c}{$l_{\infty}$} & \multicolumn{3}{c}{$l_{0}$} & \multicolumn{3}{c}{$l_{2}$} & Physical \\
				\cmidrule(r){1-1} \cmidrule(r){2-4} \cmidrule(r){5-7} \cmidrule(r){8-10} \cmidrule(r){11-11}
				SMGAA-3 &FGSM  & BIM  & PGD  & JSMA & Sparse-RS & PGD ($l_{0}$) & DeepFool & C$\&$W & SVA ($l_{2}$)&LAVAN \\
				25.7  & 1.25 & 1.26 & 1.27 & 34.5 & 27.8 & 44.3 & 5.1 & 0.7 & 38.3 & 18.4 \\ 
				\bottomrule[1.5pt]
			\end{tabular}\label{infand2}
			\begin{tablenotes}
				\item Note: The overall highest fooling rates highlighted in \textbf{bold}, and the best results in each comparison of the proposed method, $l_{\infty}$, and $l_{0}$ attacks are are \underline{underlined}. The JSMA algorithm was only applied to AConvNet because our experimental environment failed in satisfying its GPU memory requirement for $224^{2}$ input.
			\end{tablenotes}
		\end{threeparttable}
	\end{table*}%
	
	\begin{table}[tbp] 
		\renewcommand{\arraystretch}{1.2}
		\centering
		\caption{Constraints (radii) $\epsilon$ for different types of attacks in the comparative experiments, other parameter setting is listed in Table \ref{comparesetting2}}
		\label{comparesetting}
		\begin{threeparttable}
			\begin{tabular}{cccccccccc}
				\toprule[1.5pt] 
				& \multicolumn{3}{c}{$N=1$} & \multicolumn{3}{c}{$N=2$} &\multicolumn{3}{c}{$N=3$}  \\
				\cmidrule(r){2-4} \cmidrule(r){5-7} \cmidrule(r){8-10} 
				\diagbox[width=1.1cm, height=0.6cm]{$s$}{$p$}& $\infty$ & 0 & 2 & $\infty$ & 0 & 2 & $\infty$ & 0 & 2\\
				\midrule
				88   & $0.82$ & 50 & 2.87 & $1.14$ & 79 & 3.28 & $1.40$  & 96 & 3.58 \\
				224 & $0.78$ & 342 & 6.98 & $1.08$ & 551 & 7.96 & $1.32$  & 678 & 8.68\\
				\bottomrule[1.5pt]
			\end{tabular}%
			\begin{tablenotes}
				\item Note: The $l_{\infty}$ constraints are $\epsilon/255$.
			\end{tablenotes}
		\end{threeparttable}
	\end{table}%
	
	Although our proposal already takes advantage of the physical relevance, in this section, we focus on comparing the proposed method to the currently studied $l_{p}$ attacks in SAR ATR, as well as the typical physical attack framework in the optical object recognition tasks. It should be as fair as possible since there has different distance measurements. Therefore, equivalent constraints were sought for the $l_{p}$ attacks. Specifically, the $l_{\infty}$-norm constraint (maximum absolute value for every pixel) was set as the average $l_{1}$-norm perturbation caused by SMGAA-$N$ attack, that is, $\mathbb{E}_{\bm{\delta}_{N} \sim \mathcal{X}_{N}} \left[  \|\bm{\delta}_{N}\|_{1} / (s^{2}) \right]$, where $\mathcal{X}_{N}$ represents the adversarial set generated by SMGAA-$N$ for all the models. The $l_{0}$-norm constraint (maximum available pixels to perturb) was set to the average number of pixels that contain 99$\%$ energy of $N$ adversarial scatterers (calculated by $l_{2}$-norm). 
	Table \ref{comparesetting} summarizes the corresponding constraints for different input sizes $s$, as well as values of $p$ and $N$. For fair comparison, we set no constraint for the physical attack LAVAN \cite{karmon2018lavan}. Detailed parameter settings of all the studied attacks can be found in Table \ref{comparesetting2}.
	
		\begin{table*}[tbp] 
		\renewcommand{\arraystretch}{1.2} 
		\centering
		\caption{Average fooling rates ($\%$) against eight DNNs achieved by different attack algorithms with and without various extra perturbations. The extra interference includes additive white noise ($\sigma = 10^{-2}$), Gaussian filtering ($\sigma = 1$), median filtering, and resizing (bilinear interpolation, $s^{2} \rightarrow 88^{2} \rightarrow  s^{2}$). The kernel size of filters was respectively set to $3\!\times\!3$, $7\!\times\!7$ for $s=88$ and 224}
		\label{pertrobust}
		\begin{threeparttable}
			\begin{tabular}{cccccccccc}
				\toprule[1.5pt]
				&  \textbf{Proposed} & \multicolumn{3}{c}{$l_{\infty}$} & \multicolumn{3}{c}{$l_{0}$} & \multicolumn{2}{c}{$l_{2}$ $\&$ physical} \\ 
				\cmidrule(r){2-2} \cmidrule(r){3-5} \cmidrule(r){6-8} \cmidrule(r){9-10}
				Attack & SMGAA-1  & FGSM \cite{harnessing2015goodfellow} & BIM \cite{kurakin2016adversarial} & PGD \cite{madry2018towards} & JSMA \cite{jsma} & Sparse-RS \cite{croce2020sparse} & PGD ($l_{0}$) \cite{croce2019sparse} & SVA ($l_{2}$) \cite{sva} & DeepFool \cite{moosavi2016deepfool}  \\
				
				No Interference & 53.7  &  32.2 & 40.8 & 42.4 & 53.3 &55.9  & 65.6  &   80.0 & 96.1   \\ \midrule
				Add. Noise& \textbf{53.7}  & 23.6  & 27.4 & 30.2 & 51.4 & 52.5 & 64.9 & \underline{79.5}  & 30.5  \\
				Gau. Filtering& \underline{43.0}  &  6.5 & 5.9 & 5.7 & 39.7 & 2.4 & 19.4 & \textbf{77.0} & 9.0  \\
				Med. Filtering& \underline{40.1}  &  3.9 & 2.6 & 3.1 & 6.3 & 0.1 & 0.2  & \textbf{69.2} & 4.4  \\
				Resizing &  \textbf{53.7} & 4.7  & 4.0 & 3.8 & \underline{53.3} & 4.5 & 21.5 &  76.7&13.8  \\  \midrule
				Attack & SMGAA-2  & FGSM \cite{harnessing2015goodfellow} & BIM \cite{kurakin2016adversarial} & PGD \cite{madry2018towards} & JSMA \cite{jsma} & Sparse-RS \cite{croce2020sparse} & PGD ($l_{0}$) \cite{croce2019sparse} & SVA ($l_{2}$) \cite{sva} & C$\&$W \cite{carlini2017towards}  \\
				No Interference & 77.1  &  40.7 & 48.1 & 49.1 & 66.9 & 61.5 & 76.2 &  83.5 & 99.8 \\ \midrule
				Add. Noise& \textbf{77.0}  & 34.4  & 40.1 & 42.4 & 63.8 & 58.1 & 75.5 & \underline{83.1} & 19.3 \\
				Gau. Filtering& \underline{62.6}  & 10.4  & 9.4 & 8.9 & 44.8 & 4.1 & 33.1 & \textbf{81.1} & 0.2 \\
				Med. Filtering&  \underline{56.5} & 6.9  & 4.7 & 5.3 & 10.9 & 0.1 & 0.2 & \textbf{74.2} & 1.1 \\
				Resizing & \textbf{77.1}  &  7.7 & 6.5 & 6.3 & 66.9 & 8.1  & 34.7 &  \underline{80.6} & 12.7 \\  \midrule
				Attack & SMGAA-3 & FGSM \cite{harnessing2015goodfellow} & BIM \cite{kurakin2016adversarial} & PGD \cite{madry2018towards} & JSMA \cite{jsma} & Sparse-RS \cite{croce2020sparse} & PGD ($l_{0}$) \cite{croce2019sparse} & SVA ($l_{2}$) \cite{sva} & LAVAN \cite{karmon2018lavan} \\
				No Interference & 87.2  &  46.3 & 52.8 & 53.8 &  71.7& 63.4 & 80.4 & 85.6 & 38.9 \\ 	\midrule
				Add. Noise& \textbf{86.8}  & 41.1  & 47.4 & 49.4 & 68.0 & 59.9 & 80.0 & \underline{85.1} & 37.7 \\
				Gau. Filtering& \underline{70.4}  & 13.9  & 12.6 & 12.1 & 48.9 & 5.7 & 39.8 & \textbf{83.4} & 2.7 \\
				Med. Filtering& \underline{62.7}  & 9.7  & 6.6 & 7.6 & 11.9 & 0.1 & 0.2 & \textbf{77.0} & 2.2 \\
				Resizing &\textbf{87.2}  & 10.3 & 8.9 & 8.5 & 71.7 & 10.0 & 41.3 & \underline{83.1} & 3.0 \\
				\bottomrule[1.5pt]
			\end{tabular} 
			\begin{tablenotes}
				\item Note: The most and second robust results are respectively \textbf{bolded} and \underline{underlined}.
			\end{tablenotes}
		\end{threeparttable}
	\end{table*}%
	
	\begin{table}[tbp] 
		\renewcommand{\arraystretch}{1.2}
		\centering
		\caption{Parameter settings and perturbation scale for different attack algorithms}
		\label{comparesetting2}
		\begin{threeparttable}
			\begin{tabular}{ccc}
				\toprule[1.5pt] 
				Attacks &Parameters & Scale \\ \midrule
				SMGAA & - & $88^{2}$ \\ 
				FGSM\tnote{*} \cite{harnessing2015goodfellow} & $eps=\epsilon$ & $s^{2}$ \\
				BIM\tnote{*} \cite{kurakin2016adversarial} & $eps=\epsilon, steps=10, \alpha=\epsilon/10$  & $s^{2}$\\
				PGD\tnote{*} \cite{madry2018towards} & \makecell{ $eps=\epsilon, steps=20, \alpha=\epsilon/10$\\$ random\_start=True$} &$s^{2}$ \\
				JSMA\tnote{\#} \cite{jsma} & \makecell{$clip\_min=0, clip\_max=1,\theta=1,\gamma=\epsilon/s^{2}$\\$y^{target}=(y^{gt}+1)\mod10$} & $s^{2}$ \\
				Sparse-RS\tnote{\$} \cite{croce2020sparse} & \makecell{ $eps=\epsilon, n\_queries=1000,p\_init=.8$\\$targeted=False,constant\_schedule=True$}& $s^{2}$ \\
				PGD ($l_{0}$)\tnote{+} \cite{croce2019sparse}  &\makecell{$restarts=1,steps=100,stepsize=30000/255$\\$\kappa=-1,sparsity=\epsilon$} &$s^{2}$ \\
				SVA ($l_{2}$) \cite{sva} & \makecell{$\epsilon=\epsilon,\alpha=\epsilon/10,\bm{W}=Gassuian(15^{2})$\\$\bm{\delta}=\bm{\delta}+\alpha\cdot\frac{\bm{m}\odot\bm{W}*\nabla_{x}\mathcal{L}(\bm{x}+\bm{\delta},y^{\text{gt}})}{\|\bm{m}\odot\bm{W}*\nabla_{x}\mathcal{L}(\bm{x}+\bm{\delta},y^{\text{gt}})\|_{2}}$} & $s^{2}$ \\
				DeepFool\tnote{*} \cite{moosavi2016deepfool} & $steps=50,overshoot=0.02$ &$s^{2}$ \\
				C$\&$W\tnote{*} \cite{carlini2017towards} &  $steps=1000,c=1,\kappa=0,lr=0.01$&$s^{2}$ \\
				LAVAN\tnote{$^{\circ}$}  \cite{karmon2018lavan} & \makecell{$iter=100,max-min=0.135s$\\
					$\bm{\delta}=\bm{\delta}+5/255\cdot\operatorname{sign}(\frac{\partial\ell\left(\vec{y}, y_{\text {gt}}\right)}{\partial x})$} & $s^{2}$ \\
				\bottomrule[1.5pt]
			\end{tabular}%
			\begin{tablenotes}
				\item Note: This table follows the denotations in the open source implementations or original paper.
				\item \tnote{*} \url{https://github.com/Harry24k/adversarial-attacks-pytorch}
				\item \tnote{\#} \url{https://github.com/BorealisAI/advertorch}
				\item \tnote{\$} \url{https://github.com/		 	fra31/sparse-rs}
				\item \tnote{+} \url{https://github.com/fra31/
					sparse-imperceivable-attacks}
				\item \tnote{$^{\circ}$}  \url{https://github.com/ethan-iai/LaVan-Pytorch}
			\end{tablenotes}
		\end{threeparttable}
	\end{table}%

	The fooling rates and average transfer rates (the mean value of the matrix in Fig. \ref{cf}) of all the studied attack algorithms are reported in Table \ref{infand2}, from which we can conclude the following observations. 
	Firstly, the SMGAA can achieve competitive attack performance although constrained by the extremely low-dimensional ASCM and outperform the physical attack.
	Secondly, our proposal can achieve relatively universal attack performance against all the models than the $l_{\infty}$ and $l_{0}$ attacks. The performances of the later attacks are shown highly dependent on the target models. For example, they can achieve high fooling rates to some of the studied models (like DenseNet121 and ShuffleNetV2) but extremely low fooling rates for others (like VGG11, AConvNet, and SqueezeNet). 
	Thirdly, whereas $l_{0}$ attacks are capable of fooling the models through modifying limited pixels, the resulting sparse perturbations are visually abrupt and have no physical relevance to the SAR target images (as shown in Fig. \ref{sparsecomp}). Lastly, the $l_{2}$ attacks can effectively fool the models with finite perturbations and have a universal attack performance to different DNN structures. It should be noted that the $l_{2}$ variant of the SVA \cite{sva} attack accumulates perturbations to maximize utilization of the given constraints, while DeepFool \cite{moosavi2016deepfool} and C\&W \cite{carlini2017towards} pursue the smallest perturbation.
	
	\begin{figure}[tbp]
		\centering
		\includegraphics[width=0.95\linewidth]{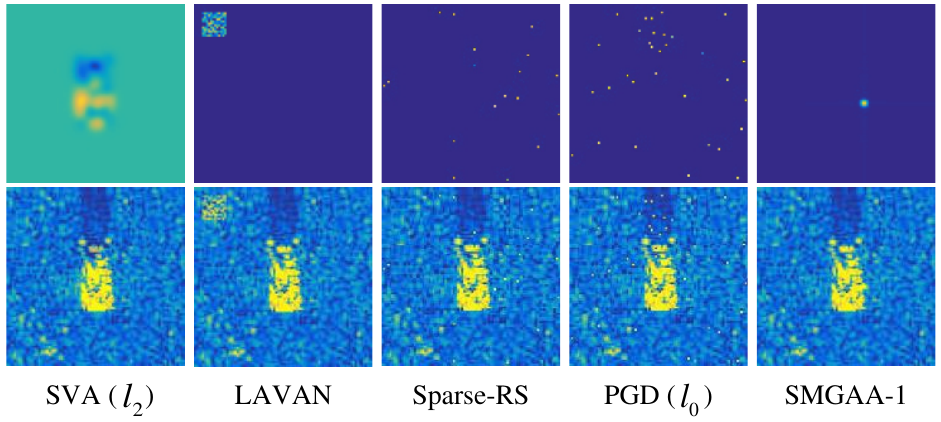}
		\caption{Adversarial examples and corresponding perturbations of the AConvNet generated by the compared methods.}
		\label{sparsecomp}
		\vspace{-15pt}
	\end{figure}

	We then investigate the robustness of these attacks. Various interference was further applied to test the perturbations' robustness. Table \ref{pertrobust} elaborates the details and results which show that the minimal perturbations generated by DeepFool and C\&W attacks are the most non-robust. The results align with the widely held belief that extremely small perturbations would easily be disrupted by noise or transformations. We can also observe that the SMGAA and the local accumulation-based attack (SVA ($l_{2}$)) exhibit the  robustness in the face of every interference, with distinctly higher remained average fooling rates than other algorithms. It should be pointed out that the resize operation was considered assuming that the scaled adversarial examples will be implemented to their original size. In this context, our algorithm just generates $88^{2}$ adversarial images and is free of resizing.  
	\begin{table*}[htbp]
		\renewcommand{\arraystretch}{1.2}
		\centering
		\caption{Fooling rates ($\%$) against eight DNN models with achieved by SMGAA-1 with different initialization locations}
		\begin{threeparttable}
			\begin{tabular}{cccccccccc}
				\toprule[1.5pt]
				\makecell{Initialization\\ of location} 	& AlexNet  & VGG11  & ResNet50  & DenseNet121& MobileNetV2  & AConvNet   & ShuffleNetV2& SqueezeNet & \textbf{Average}\\
				\midrule
				Whole Image & 48 & 33 & 37 & 23 & 21 & 32 & 33 & 45 & 34.0	\\
				Background & 28 & 21 & 30 & 8 & 14 & 19  & 18 & 29 & 20.9	\\
				Target$\&$Shadow  & \textbf{57} & \textbf{45} & \textbf{48} & \textbf{33} & \textbf{34} & \textbf{48} & \textbf{45} & \textbf{55}  & \textbf{45.6} \\
				\bottomrule[1.5pt]
			\end{tabular}\label{abation2}
			\begin{tablenotes}
				\item Note: Best result in each \textit{column} is highlighted in \textbf{bold}.
			\end{tablenotes}
		\end{threeparttable}
	\end{table*}%

		\begin{figure}[tbp]
		\centering
		\includegraphics[width=0.95\hsize]{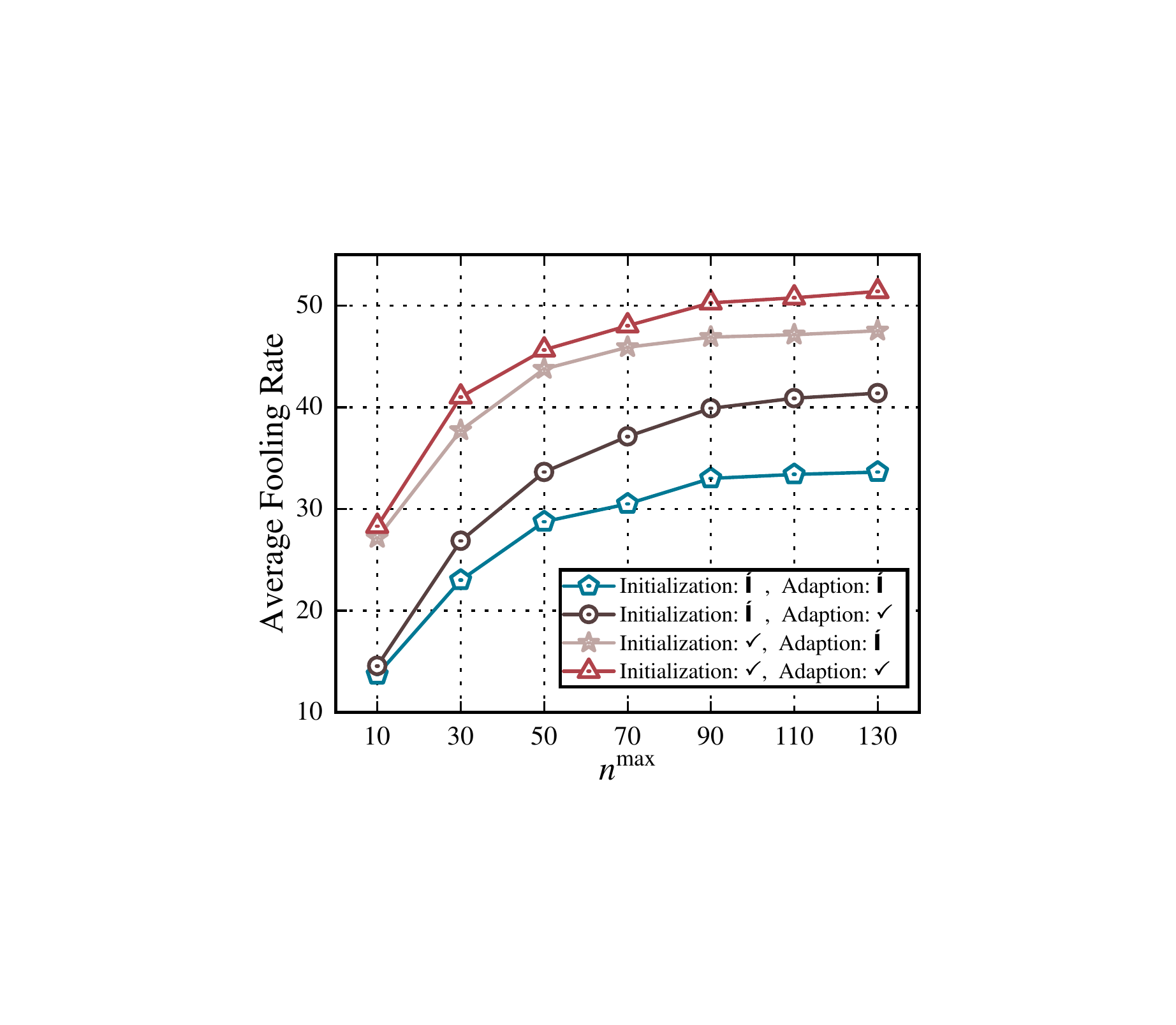}
		\caption{Average fooling rates ($\%$) against eight DNN models achieved by SMGAA-1 with different optimization strategies and maximum iterations.}
		\label{method}
	\end{figure}

	\subsection{Ablation Study} \label{ablation}
	In this section, we investigate how effective is the proposed generation algorithm, and its sensitivity to the main ASCM attributes such as the scattering type, amplitude, and location. For efficiency, we selected a 100-image subset to perform the SMGAA-3 attack with $n^{\text{max}}$ and $B_{N}$ both set to 50. Other setting was consistent with the afore-listed setting in Table \ref{paras} unless otherwise stated.
	
	\subsubsection{Efficacy of the Proposed Optimization Strategies}
	
	Fig. \ref{method} depicts the attack performance (average fooling rate on eight models) of the different combinations of the proposed initialization and adaption strategies. When $n^{\text{max}}$ varies from 10 to 130, all the three conditions exhibit stable improvements, and the initialization strategy based attack perform eye-catching better than the vanilla attack. The reason is that the optimization is extremely prone to be stuck in local optima subject to the location modality, thus, a good initial location can significantly promote the whole process. The adaption is also helpful to find an available solution in limited steps. Moreover, the proposed two strategies are compatible to bring more advances to the optimization.

		\begin{figure}[tbp]
		\centering
		\includegraphics[width=0.9\linewidth]{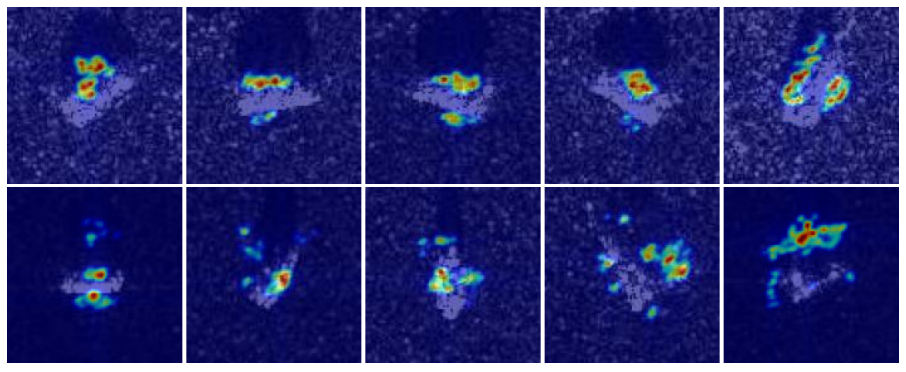}
		\caption{Heatmaps of the vulnerable region of SAR target images, which are the superposition of a hundred attacks to AConvNet (SMGAA-1 with $v_{\text{th}} = 0.2$).}
		\label{location}
	\end{figure}
	\begin{figure}[tbp]
		\centering
		\includegraphics[width=0.95\hsize]{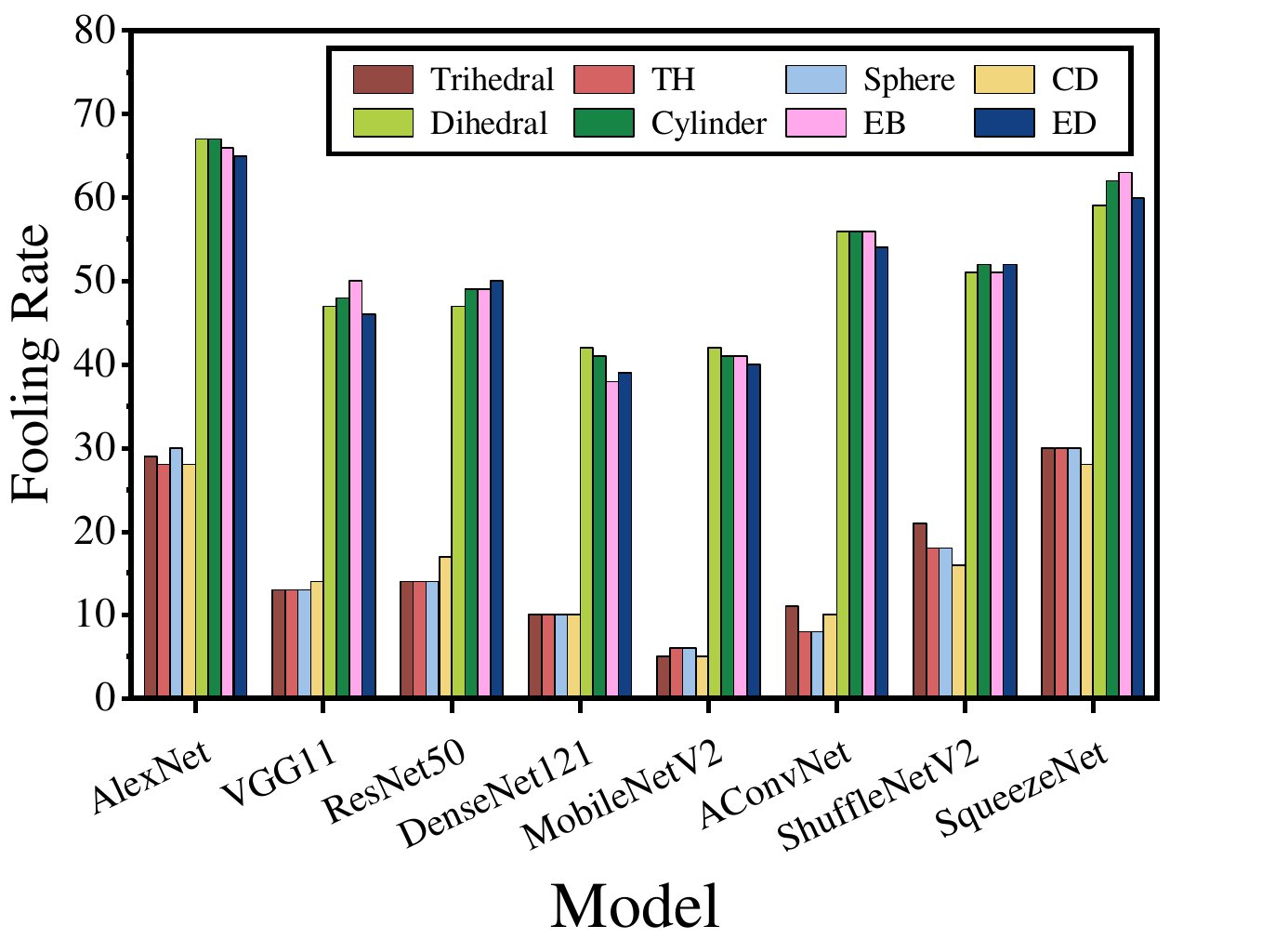}
		\caption{Fooling rates ($\%$) against eight DNN models achieved by SMGAA-1 with different scattering types.}
		\label{type}
	\end{figure}

	\subsubsection{Influence of Location}
	Further, we investigate the influence of the initial location, with results summarized in Table \ref{abation2}. Our strategy that initializes the candidate scatterers in the target and shadow region succeeds in achieving the highest fooling rates on each of the models. It can be observed that the scatterers initialized in background clutter perform poorly than the uniformly random initial location on the whole image. The results support the intuition that the target and shadow parts give rise to most of the informative features due to they carry the structural and scattering information. 
	
	In Fig. \ref{location}, we show the sensitive region of the SAR target images by performing a hundred attacks (SMGAA-1 with $v_{\text{th}} = 0.2$) for each image and superposing the scatterers' location. In the figure, the brighter regions are more vulnerable to the perturbations in the form of scatterers. The results show that the vulnerable regions are mainly the electromagnetic shadowing parts both on the target and background, as well as the sides of the target. The result leads to the assumption that the DNN may implicitly learn the structural information from the training data and is highly sensitive to structural deformations.

	\begin{table*}[htbp]
	\renewcommand{\arraystretch}{1.2}
	\centering
	\caption{Fooling rates ($\%$) against eight DNN models with achieved by SMGAA-1 with different amplitude constraints}
	\begin{tabular}{cccccccccc}
		\toprule[1.5pt]
		\makecell{Maximum \\ amplitudes (dB)}	& AlexNet  & VGG11  & ResNet50  & DenseNet121& MobileNetV2  & AConvNet   & ShuffleNetV2& SqueezeNet &  \textbf{Average} \\
		\midrule
		0 & 57 & 45 & 48 & 33 & 34 & 48 & 45 & 55 &	 45.6 \\
		0.5 & 75 & 57 & 66 & 66 & 60 & 75 & 76 & 77 & 69.0	\\
		1.0 & 81 & 65 & 78 & 77 &  75 & 76 & 82 & 81 &76.9	\\
		1.5 & 81 & 68 & 81 & 80 & 77 & 78 & 86 & 83 &79.3	\\
		\bottomrule[1.5pt]
	\end{tabular}\label{abation4}
\end{table*}%
\begin{figure*}[htbp]
	\centering
	\includegraphics[width=1.0\linewidth]{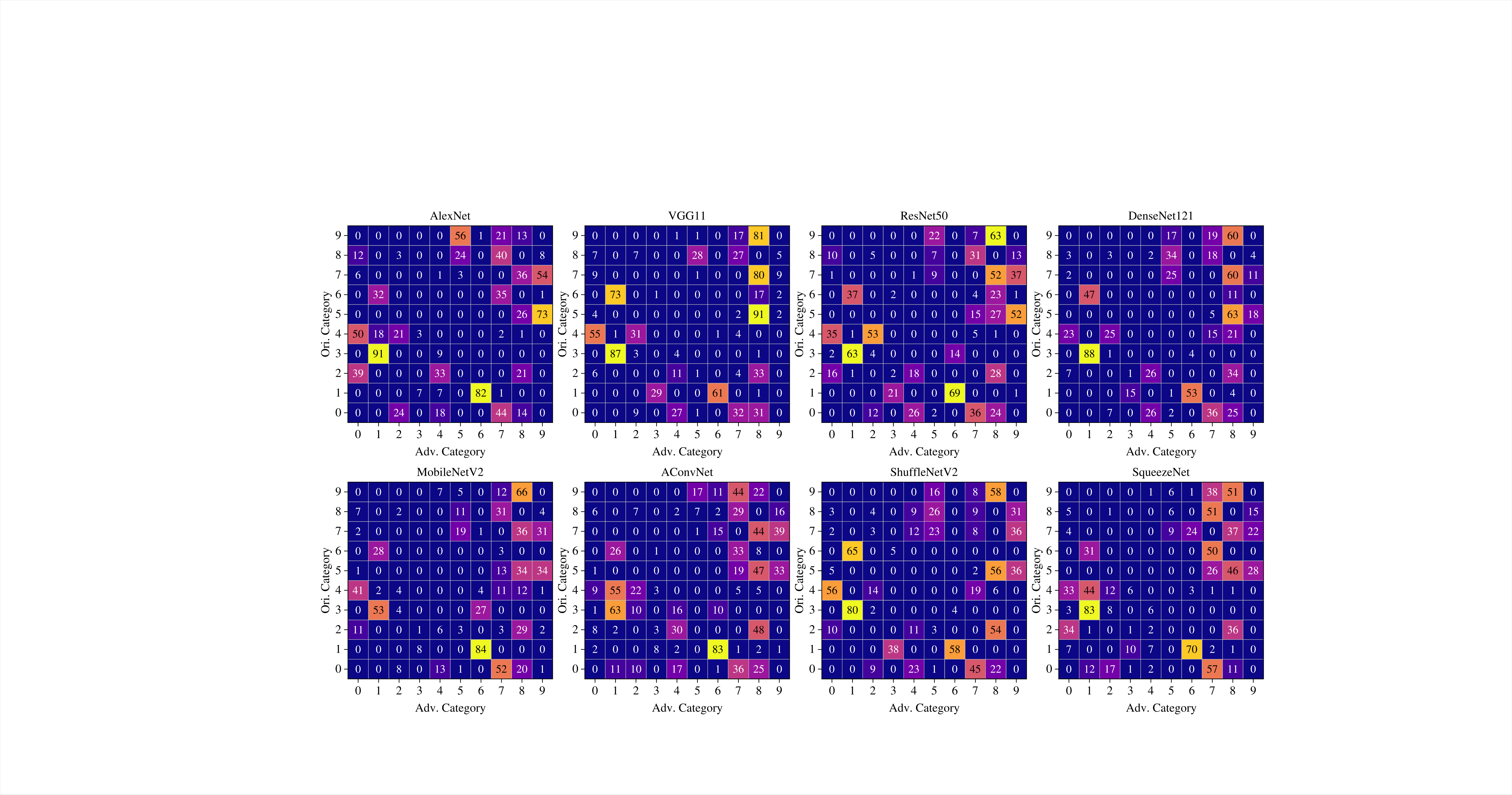}
	\caption{Distribution of adversarial categories of the SMGAA-3 attack. The \textit{Ori. Category} refers to the ground truth category of targets, \textit{Adv. Category} denotes the misclassified category of the corresponding adversarial examples. The number from 0 to 9 indicates respectively the following classes: 2S1, BMP2, BRDM2, BTR70, BTR60, D7, T72, T62, ZIL131, ZSU234.}
	\label{fig4}
\end{figure*}
	\subsubsection{Effect of Scattering Type}
	Fig. \ref{type} summarizes the fooling rates achieved by different scattering types. In the experiment, each batch of candidate scatterers was initialized by a single scattering type. There is an overview that the distributed scattering mechanism is more aggressive due to it alters more pixels of the resulted image. The average fooling rate of distributed scatterers is 57.5$\%$, higher than the one of located scatterers by 35.1$\%$, and the fooling rates within the two groups are not significantly different. It suggests that no careful configuration of frequency dependence is required when initializing the perturbation.
	
		\subsubsection{Effect of Amplitude} \label{supperession}
	The SAR images are usually maximum-normalized in the end of imaging process, which allows a locally strong scatterer to suppress the whole target response. We take this suppression as a weakness the SMGAA can leverage. Table \ref{abation4} exhibits the fooling rates when the amplitude of the adversarial scatterer (relative to the largest amplitude of the target) ranges from 0 to 1.5 (dB). It can be seen that the average fooling rate against the eight models increases by 22.5$\%$, 30.4$\%$, and 32.8$\%$ at 0.5, 1, and 1.5 dB, respectively. Note that this suppression differs from the conventional suppression jamming. It mainly works by strengthening the adversarial features and destructing original features in high-dimensional space. It also may take effect by causing the statistical excursion. Practically, all the studied models are sensitive to this mild suppression, suggesting carefully distinguishing the strong scattering centers in the circumstance that exists potential adversarial risk.

	\subsection{Discussion on the Attack Selectivity of MSTAR Data} \label{discussion}
	It has shown that the MSTAR data has attack selectivity \cite{sarAAempirical2021,sarAAexperience2020}, that is, the misclassified categories of the adversarial examples are concentrated and exhibit high consistency across different attack algorithms and victim models. For instance, most of the adversarial examples of D7 would be recognized as ZIL131, ZSU234, and T62 for various attack algorithms and victim models. The distribution of adversarial categories caused by the proposed method is reported in Fig. \ref{fig4}. Each element in the sub-figure denotes the original-adversarial (row to column) category pairs under the SMGAA-3 attack. It illustrates that each original category is mainly misclassified to 2$\sim$5 adversarial categories, which means different data points of the same category share the similar vulnerable categories. Moreover, there is a certain symmetry across all the matrices, and all the matrices exhibit a certain similarity, suggesting the attack selectivity is not only related to the model structure and attack method, but as well as to the distribution of the MSTAR data, and more fundamentally, the targets themselves. Since it is difficult for a non-expert to distinguish the details of the SAR targets, the adversarial attack may be utilized as an analytical tool to learn the inter-categories similarity and its imaging representation.

	\section{Conclusion}\label{conclusion}
	In this article, we have reported the framework, algorithm, results, as well as the analysis of the SMGAA. To generate more feasible and physically informative adversarial perturbations for DNN-based SAR ATR, the ASCM was introduced as an electromagnetic constraint for the attack process, ensuring the physical feasibility of the resulted adversarial perturbations. It is hard to efficiently find solutions on the extremely intricate loss surface of DNNs. To achieve that, several customized, simple, yet effective strategies were proposed to facilitate the general gradient-based optimization, showing considerable improvements. Compared with the currently studied $l_{p}$ attacks in this field, our proposal exhibited universal robustness in attacking various DNN structures and against diverse perturbations. 
	More importantly, the proposed method is promising to serve as a surrogate to construct robust ATR models that are defensive against malicious scatterers. In addition, the inter-category similarity and the sensitive region of SAR target images were reported, which are also potential to design defense methods.
	
	For future work, we plan to validate the proposed method in the real world and and exploit the proposed method for more powerful robust model training. We would also like to improve the transferability of our method to better describe the non-cooperative attack threats.

	\section*{Acknowledgments}
	This work was supported partially by the National Key Research and Development Program of China under Grant 2021YFB3100800, National Natural Science Foundation of China under Grant 61921001, 62022091, and the Changsha Outstanding Innovative Youth Training Program under Grant kq2107002.
	
	\bibliographystyle{IEEEtran}     
	\bibliography{refediffasc}

\end{document}